\documentclass{llncs}

\usepackage{algorithm, float, graphicx, mathtools, multirow, array, makecell, hhline, tabularx, amssymb, amsmath, subcaption}

\usepackage{xcolor}
\usepackage[normalem]{ulem}
\usepackage{graphicx}
\usepackage{algorithm}
\usepackage{algpseudocode}
\usepackage{pdflscape,soul}
\usepackage{xr} % External reference package

\usepackage{hyperref}
\usepackage{tikz}
\usepackage{booktabs}
\usepackage{todonotes}
\usepackage{etoolbox}
\usepackage{siunitx}
\hypersetup{urlcolor=blue}
\usepackage{xcolor}
\newcolumntype{C}[1]{>{\centering\let\newline\\\arraybackslash}p{#1}}
\newcolumntype{L}[1]{>{\raggedleft\let\newline\\\arraybackslash}p{#1}}
\newcolumntype{R}[1]{>{\raggedright\let\newline\\\arraybackslash}p{#1}}

\usetikzlibrary{shapes.geometric, arrows}
\tikzstyle{process} = [rectangle, minimum width=3cm, minimum height=1cm, text centered, draw=black, fill=white!30]
\tikzstyle{arrow} = [thick,->,>=stealth]

\usepackage{xcolor}
\usepackage{graphicx}

\begin{document}

\author{Niki van Stein\inst{1}\orcidID{0000-0002-0013-7969} \and
Sarah L. Thomson \inst{2}\orcidID{0000-0001-6971-7817} \and
Anna V. Kononova\inst{1}\orcidID{0000-0002-4138-7024}
}
\authorrunning{N. van Stein et al.}
\institute{
LIACS, Leiden University, Einsteinweg 55, 2333 CC Leiden, Netherlands\\
\email{\{n.van.stein,a.kononova\}@liacs.leidenuniv.nl} \and
Edinburgh Napier University, United Kingdom\\
\email{s.thomson4@napier.ac.uk}
}
%

%\title{Affine Trajectories \\and the Interplay with Structural Bias} % Sarah's
%\title{Interplay of Performance with Structural Bias along Affine Trajectories} % Anna's suggestion
% Nikis suggestion:
\title{A Deep Dive into Effects of Structural Bias on CMA-ES Performance along Affine Trajectories}
% sarah:
%\title{Interplay of structural bias with structural bias along affine trajectories}
\maketitle
\begin{abstract}
To guide the design of better iterative optimisation heuristics, it is imperative to  understand how inherent structural biases within algorithm components affect the performance on a wide variety of search landscapes.
This study explores the impact of structural bias in the modular Covariance Matrix Adaptation Evolution Strategy (modCMA), focusing on the roles of various modulars within the algorithm. Through an extensive investigation involving $435\,456$ configurations of modCMA, we identified key modules that significantly influence structural bias of various classes. Our analysis utilized the Deep-BIAS toolbox for structural bias detection and classification, complemented by SHAP analysis for quantifying module contributions. The performance of these configurations was tested on a sequence of affine-recombined functions, maintaining fixed optimum locations while gradually varying the landscape features. Our results demonstrate an interplay between module-induced structural bias and algorithm performance across different landscape characteristics.
\end{abstract}

\setcounter{footnote}{0} 
%--------------------------------------------------------------------

\section{Introduction}
In light of the rapid advancement of the field of heuristic black-box optimisation~\cite{thirty_years}, a remarkable array of algorithms is now available to practitioners. The behaviour of most of these algorithms strongly depends on the settings of numerous hyperparameters, exploding the number of options further and making the choice of a well-performing algorithm configuration for a specific (real-world) problem even harder. 

And yet, we still don't understand these algorithms well enough. One thing we can do is screen (a family of) algorithms or algorithm configurations against some unwanted characteristics. Although it is unrealistic to examine all settings across all characteristics, initial efforts are essential. Such screening is expensive but it not only helps eliminate ineffective configurations but also aids in elucidating the internal dynamics that impedes performance. Modular algorithm designs~\cite{VermettenModDE2023,de2021tuning}, where each operator option can be selected independently of the choice for other operators, are particularly well-suited for such analyses.

While in many cases, the unwanted characteristics only manifest within specific function landscapes (and the correspondence between these landscapes and characteristics can be unknown), it is hypothesised that at least some characteristics can be assessed in general. One such aspect that has not been investigated sufficiently is structural bias (SB) ~\cite{Kononova2015} and especially its precise causes and influence on the algorithm's performance. SB is the algorithm’s inherent limitation in locating optima in certain regions of the domain independently of the objective function’s landscape. It stems from the iterative application of a limited set of algorithm operators and their interplay~\cite{Kononova2015}. While some families of algorithms have been screened to some extent~\cite{vanstein2024explainable,VermettenModDE2023}, no clear performance implications have been established so far. Unfortunately, even though such screening is done for very large algorithm configuration spaces, it necessarily remains limited due to the need to discretise numerous continuous hyperparameters, thus potentially overlooking some interactions. This paper is no exception, however, its experimental design is structured to be as comprehensive as computationally feasible.

\begin{figure}[!t]
    \centering
    \includegraphics[width=\textwidth, trim=5mm 5mm 5mm 5mm,clip]{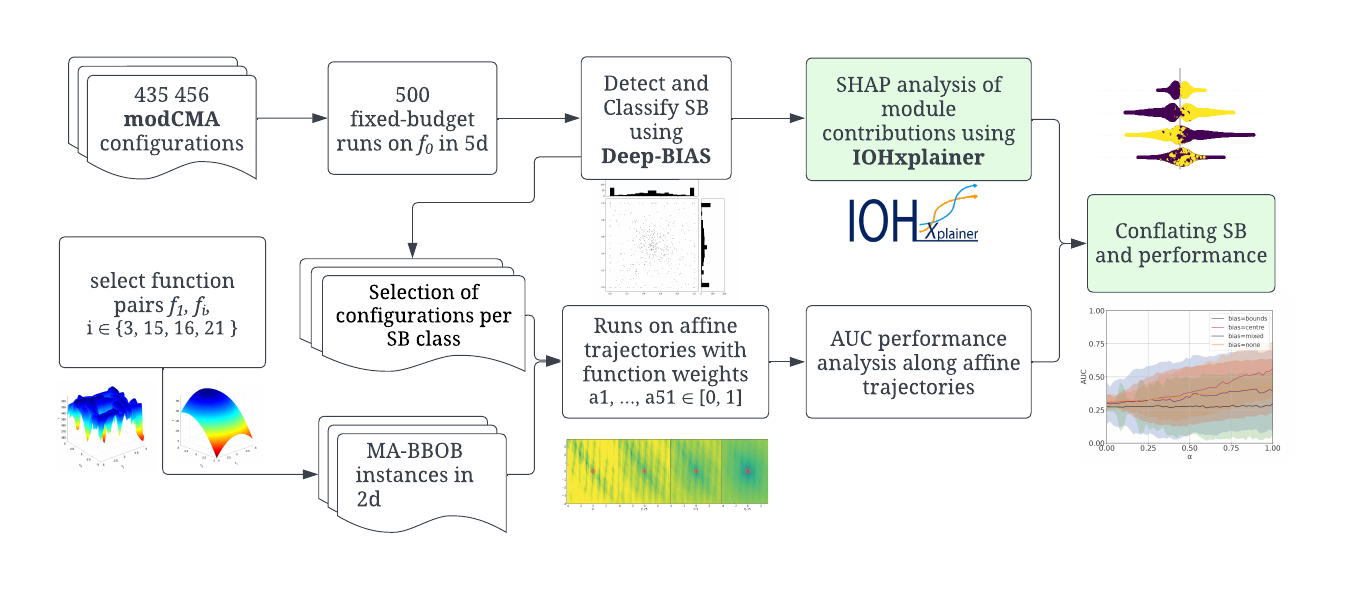}
    \caption{The summary of the overall methodology used. Full details of all steps are provided in the corresponding sections. Green blocks highlight the contributions of this paper. }\label{fig:methodology} % Anna will re-do this figure after the submission
\end{figure}

This paper aims to explore the impact of SB on algorithm performance by addressing the following questions: 1. How does the performance of structurally biased versus unbiased configurations change on sequences of functions where the landscape progressively shifts from rugged to smooth? 2. How does the location of the optima within the domain~\footnote{This paper focuses on box-constrained minimisation problems.} of these functions affect the performance depending on the class of structural bias? The complete methodology of our investigation is summarised in Figure~\ref{fig:methodology}. 

%1. How does the performance of structurally biased vs unbiased configurations from modular CMA-ES change on a smooth sequence of affine-recombined functions on which locations of optima are kept fixed? 2. How does the location of the optima affect the performance depending on the kind of structural bias?
%\sat{I have removed the third question (ELA along affine) because Diederick et al have recently done this}

%\ak{move methodology tikz to here} See Figure~\ref{fig:methodology}
%\input{tikz}
%\ak{Add drawing of our overall methodology}: (1. modCMA configuration space definition $\to$ $435\,456$ configurations run 500 \hl{?} times with fixed budget on $f_0$ in \hl{how many?} dimensions $\to$ Deep-SB test $\to$ 2. SHAP analysis of module contributions to SB per class 3. \hl{20} most biased configurations by class of SB 4. MA-BBOB instances from function pairs \emph{f1} and \emph{fi}, $i\in\{$\emph{3}, \emph{15}, \emph{16}, \emph{21}$\}$ with controlled placement of optima within the domain (at bounds, at centre, at bounds and centre \hl{?!}, uniform in $[-2,2]^d$) $\to$ affine trajectories via function weights $\alpha_1,..,\alpha_{51}\in [0,1]$) $\to$ 5. AUC performance (median, variance) along trajectories, grouped by function and SB type $\to$ conflating SB and performance

\section{Background}

% \ak{me} Background of BBOB~\cite{finck2010real}, SB~\cite{Kononova2015,Davarynejad2014}, generalised signature test~\cite{Rajwar2024}, Related work on SB~\cite{Kononova2020PPSN,VermettenDEbook2022,structcmaVermetten2022}, CMA-ES~\cite{thirty_years}, MA-BBOB~\cite{VermettenMABBOB2023,Dietrich2022} 

%\ak{1/3 page max}

\subsection{Modular CMA-ES}
Covariance Matrix Adaptation Evolution Strategy (CMA-ES) \cite{cmaes} is a robust, state-of-the-art evolutionary algorithm used for solving non-linear, non-convex optimisation problems \cite{thirty_years}. Central to its approach is the adaptation of a covariance matrix which determines the shape and scale of the search distribution, effectively learning the landscape of the problem space. This self-adaptive mechanism allows the algorithm to balance exploration of the search space with exploitation of known good regions, making it particularly effective in a wide range of practical applications, from machine learning parameter tuning to engineering design optimisation.
The Modular CMA-ES (modCMA) \cite{de2021tuning} is a Python and C++ modular implementation of the CMA-ES algorithm and many of its variants, with module options and hyper-parameters that can be switched on and off independently of each other. 
In this work we investigate the full scale of these module options, given in Table \ref{tab:cma}, leading to a total of $435\,456$ different CMA-ES configurations.

\begin{table*}[!b]
    \centering
    \caption{Considered modules of \textbf{modCMA} and their configurations.\label{tab:cma}}
    \begin{tabular}{llp{55mm}}

       \textbf{Module name} & \textbf{Shorthand} & \textbf{Domain} \\ \hline \hline
        Covariance adaptation & \texttt{covariance} &  $\{${false}, {true}$\}$ \\
        Elitism & \texttt{elitist} &  $\{${false}, {true}$\}$ \\
        Active update & \texttt{active} &  $\{${false}, {true}$\}$ \\
        Base sampler & \texttt{base\_sampler} &  $\{${Gaussian}, {Halton}, {Sobol}$\}$ \\
        Orthogonal sampling & \texttt{orthogonal} & $\{${false}, {true}$\}$\\
        Threshold convergence & \texttt{threshold} & $\{${false}, {true}$\}$\\
        Sample Sigma & \texttt{sigma} & $\{${false}, {true}$\}$\\
        Bound correction & \texttt{bound\_correction} & $\{${off}, {saturate}, mirror, COTN, toroidal, uniform$\}$\\
        Mirrored sampling & \texttt{mirrored} &   $\{${off}, {mirrored}, {mirrored pairwise}$\}$ \\
        Recombination weights & \texttt{weights\_option} &  $\{${default}, {equal}, {$\lambda$-decay}$\}$\\
        Step size adaptation & \texttt{step\_size\_adaptation} &  $\{${CSA}, {PSR}, TPA, MSR, XNES, MXNES, LPXNES$\}$ \\
        Local restarts & \texttt{local\_restart} & $\{${none}, {IPOP}, {BIPOP}$\}$ \\ 
        \hline
    \end{tabular}
\end{table*}

\subsection{Structural bias}
Structural bias in iterative optimisation heuristics \cite{Kononova2015,Davarynejad2014} refers to the tendency of certain algorithms or configurations of algorithms to favour specific regions of the search space over others, despite the absence of initial information indicating where high-quality solutions might reside. Ideally, an optimisation algorithm should explore the defined domain boundaries without preconceived preferences, allowing the data collected from sampled points to guide its progression towards areas with optimal objective function values. However, in practice, some algorithms inherently exhibit a preference, such as a bias towards the centre of the domain, which can limit their effectiveness in universally discovering the best solutions across the entire feasible domain. This phenomenon, known as \emph{structural bias}, can compromise the algorithm's ability to perform well in general situations. Detecting structural bias is hard, as the objective function and behaviour of the algorithm are always entangled. Using a special objective function ($f_0$), defined as a completely random fitness landscape, allows one to detect structural bias using statistical tests as introduced in \cite{bib:BIAS}. There are a few related works on Structural Bias that either introduced a different detection method, like the generalised signature test~\cite{Rajwar2024}, or analysed different groups of algorithms regarding SB~\cite{Kononova2020PPSN,VermettenDEbook2022,structcmaVermetten2022}.

\subsection{SHAP}
Shapley Additive Explanations (SHAP), as introduced by Lundberg et al. \cite{lundberg2017unified}, is a popular \emph{explainable AI} (XAI) method for attributing features in model predictions. SHAP quantifies the impact of a specific feature, $f$, by comparing model outputs with and without $f$. The difference in outputs, averaged across models, defines the SHAP value, which can be positive, negative, or zero, representing the feature’s marginal contribution.

However, SHAP's application to large datasets is computationally intensive. To address this, the TreeSHAP method \cite{lundberg2020local2global} employs tree-based model structures to streamline computations, using approximation techniques to enhance efficiency in scenarios with extensive feature sets.
In this work, we use this XAI method to compute the contributions of different module settings towards specific structural bias classes. This is done by training Xg-boost regression models on the one-hot-encoded algorithm configurations as input and the predicted SB class label from Deep-BIAS as output. Using these models we can approximate the SHAP values of each module option per configuration.

\section{Structural Bias Classification}

To assess the structural bias (SB) and in the end the interplay of structural bias with performance depending on landscape features and the location of the optima, the first step is to detect and classify structural bias per algorithm configuration.
We conduct a configuration sweep for modCMA \cite{de2021tuning} using the \textsc{modcma} package in Python. In this extensive analysis, we use a full grid of all of the categorical module options in modCMA, as specified in Table \ref{tab:cma}. This resulted in a total of $435\,456$ configurations. For each of these configurations the population sizes are fixed to $\mu = 5, \lambda = 20$. The analysis in this paper is broader and more in-depth than previous analysis of structural bias for modCMA \cite{structcmaVermetten2022} in the sense that it contains not just a subset of modCMA module options but the complete set of all categorical options (and a limited set of continues parameters). In addition, in this work we propose an explainable AI approach, similar to the approach used in \cite{vanstein2024explainable}, to analyze the different contributions of different module options to structural bias and to specific types of structural bias, leading to new insights. In \cite{vanstein2024explainable}, the XAI approaches was used to analyse the contributions of modules and hyper-parameters on the performance on different function landscapes, here we use it to assess the influence of modules on structural bias, and differently from the approach in \cite{vanstein2024explainable} we one-hot-encode all categorical module options to see how each option affects the structural bias individually. We also used the approach to look at second order interactions in relation with structural bias, however, these second order interactions were marginal and we therefore do not include these results in this work.

\subsection{Methodology}

For the assessment and classification of SB, we used the BIAS toolbox \cite{bib:BIAS}, available on \cite{van_stein_bas_2023_7614586}. The toolbox provides an SB detection mechanism based on the aggregation of the results of $39$ statistical tests but also a Deep-learning approach \cite{van2023deep} to detect and classify SB based on distributions of final points (found minima) of many independent runs on $f_0$. 
The three bias types we are looking at in this work are: SB towards the \textbf{centre} of the search space, towards the \textbf{bounds}, and \textbf{uniform} (no SB detected). 
SB is detected by first running an optimisation algorithm several times on the random objective function $f_0$. Here we used $100$ independent runs with $10.000$ function evaluations as budget per run to make the first classification of SB using the Deep-BIAS toolbox. Due to its speed and classification accuracy, we leveraged the Deep-BIAS model instead of the statistical methods in the BIAS toolbox. The Deep-BIAS method classifies the distribution of (in this case $100$) final best points found by the algorithm.
We do have to note that the Deep-learning model is not a perfect predictor. We therefore also verify the top $20$ configurations per SB class, used later in our experiments, by visual inspection of the final point distributions.

\begin{landscape}
\begin{figure*}[!t]
\centering
    \includegraphics[height=0.9\textwidth,trim=0mm 0mm 0mm 0mm,clip]{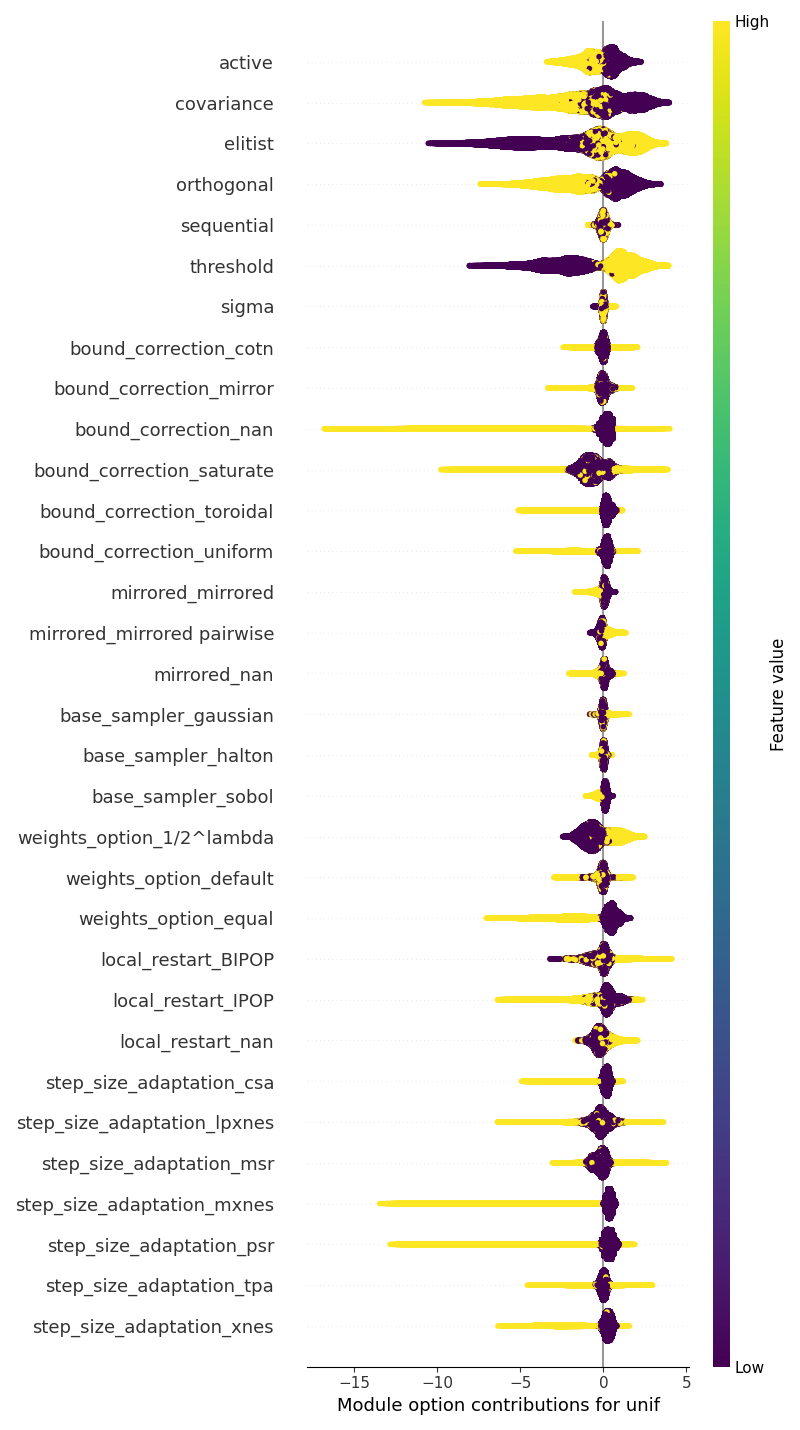}
    \includegraphics[height=0.9\textwidth,trim=0mm 0mm 0mm 0mm,clip]{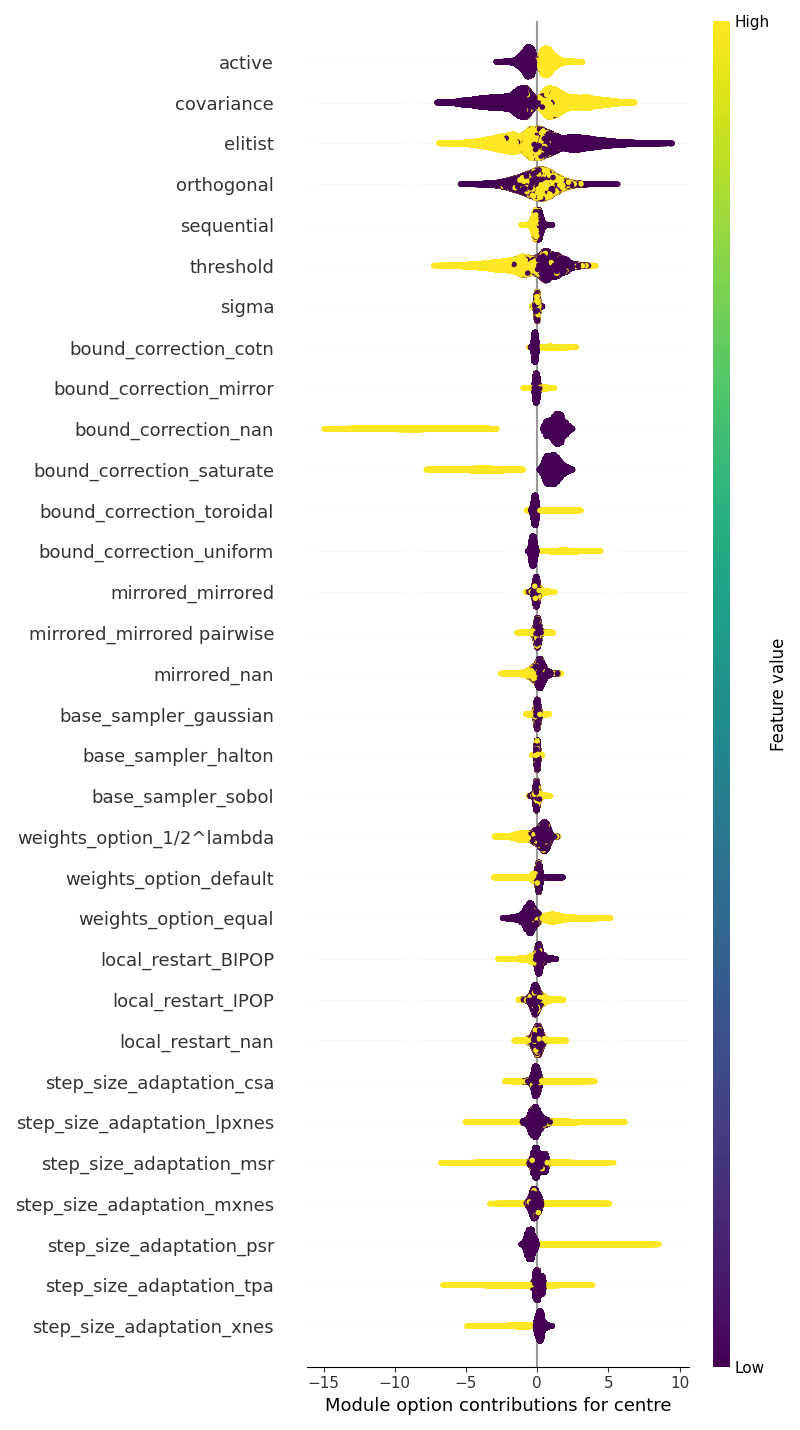}
    \includegraphics[height=0.9\textwidth,trim=0mm 0mm 0mm 0mm,clip]{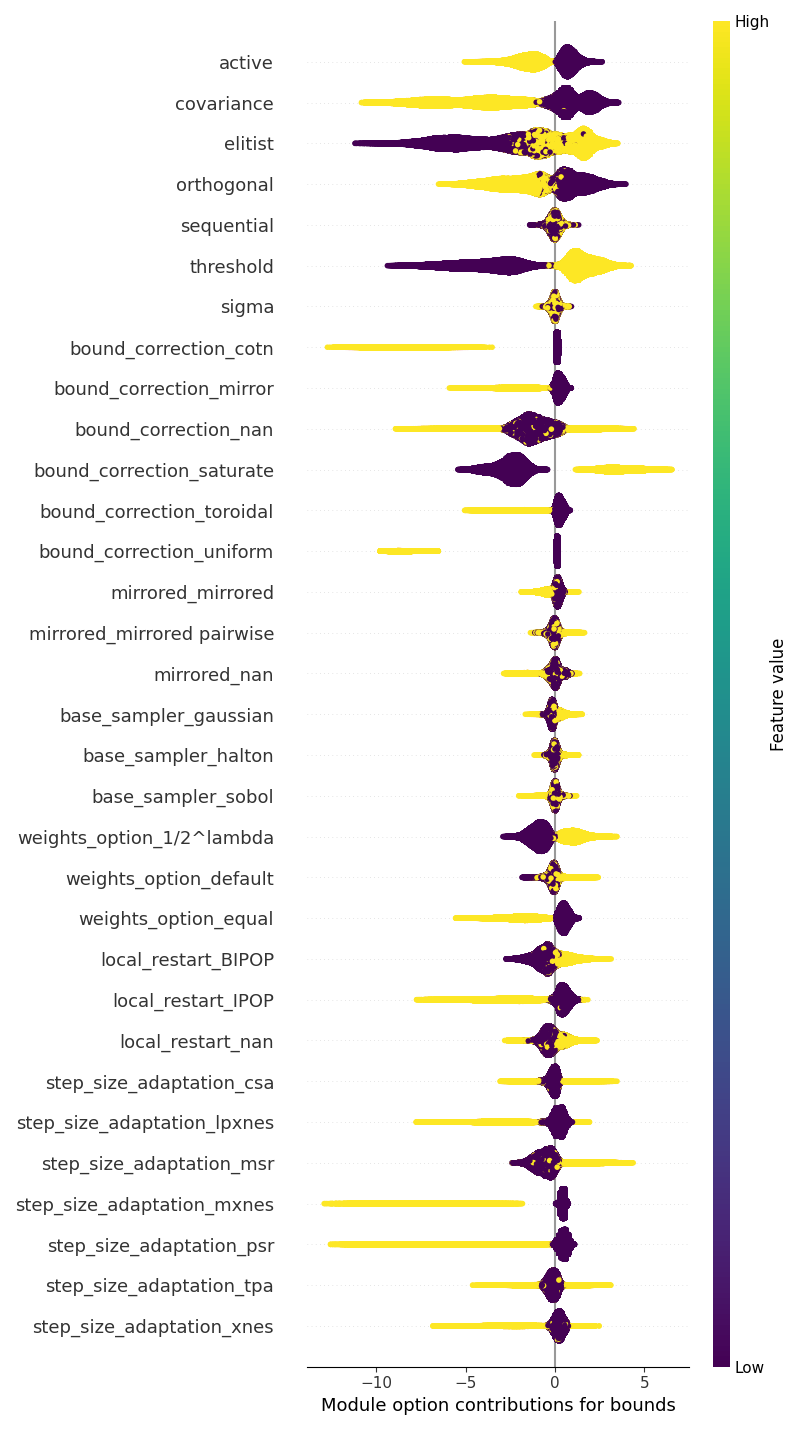}
\caption{SHAP values showing module contributions to (from left to right) no structural bias, centre bias and bounds bias classes, respectively. The baseline prediction of these classes are $0.094$, $0.559$, $0.054$ respectively, meaning that a SHAP value of 0 would result in the given baseline class probability. \label{fig:shap1}}
\end{figure*}
\end{landscape}

Once we have classified each of the configurations automatically, we can use the confidence of each SB class to calculate approximate Shapley values using the TreeSHAP algorithm \cite{lundberg2020local2global}. The calculated SHAP values for each module option per structural bias class are shown in Figure \ref{fig:shap1}. We can use these SHAP values to gain insights into which module options contribute to what kind of structural bias. 

\subsection{Module contributions to SB}

Based on the output of the Deep-BIAS package, most of the considered configurations of modCMA ($82\%$) are classified as \textbf{Centre} biased, $9\%$ as unbiased (uniform) and $5\%$ as biased towards the bounds. The remaining fraction ($3\%$) was classified as discretization bias, however after visual inspection, those configurations were mostly misclassified and should be either centre or unbiased and therefore we did not take them into account.

Given the SHAP values from Figure \ref{fig:shap1} and taking into account the base value (mean classification confidence of each class), we see a few interesting patterns. Overall, the covariance, elitism, threshold, bound correction and step size adaptation modules mostly influence the structural bias classification. We can also observe that in general, option contributions are negatively correlated between centre SB and bounds SB, in other words, when a module option causes centre SB it lowers the probability of bounds SB and vice versa. Bounds SB and uniform (no SB) seem roughly aligned except for the bound correction methods. Let us discuss the major modules involved in centre, bounds and unbiased below.

\textbf{Elitism} when turned on, reduces centre SB according to the SHAP data. Since centre SB is the majority class, Elitism seems to reduce SB in general. When looking at the inner workings of modCMA and also the objective function $f_0$, this could be explained due to the fact that with elitism the algorithm is more likely to get stuck in a (local) minimum on $f_0$ early in the optimisation run, effectively dampening the structural bias effects. Elitism by itself is however very likely not to be responsible for any structural biased behaviour.

\textbf{Threshold convergence} when turned on has a similar effect as elitism (though less profound). Again, threshold convergence is likely not causing any biased algorithm behaviour but amplifies (when turned off) or dampens (when turned on) the SB effects.

\textbf{Bound correction saturate} shows to have a large effect on bounds SB, which makes perfect sense and is in line with other research on structural bias \cite{structcmaVermetten2022}. Upon close inspection, all configurations that were classified with high confidence as bounds SB (confidence $> 0.45$), all used Saturate as the bound correction method.

\textbf{Covariance matrix adaptation} seems to play a large role in centre SB. 
In the context of a standard CMA-ES (so with the covariance module on), the search distribution is represented by a multivariate normal distribution. This distribution is characterized by its covariance matrix, which determines the shape and orientation of the points (solutions) that are sampled. Geometrically, the shape of this distribution resembles a hyper-ellipse. The search space, on the other hand, is typically a hyper-cube. This causes a mismatch in shapes being explored, likely leading to a structural bias towards the centre of the search space. The hyper-ellipse will naturally avoid sampling close to the edges and especially the corners of the hyper-cube because these areas are outside the maximum reach of the distribution whose radius is limited to the smaller distance from the center to an edge, rather than to a corner.
This effect is amplified as the dimensionality of the space increases. In higher dimensions, the corners of the hyper-cube are exponentially further away from the centre compared to the edges. Thus, a spherical sampling distribution centred in the hyper-cube will leave vast regions in the corners significantly undersampled. This results in a higher concentration of sample points towards the centre of the search space, and relatively fewer near the boundaries and corners. It could potentially lead to suboptimal exploration of the search space, especially if the global optimum lies near the boundaries or corners of the domain. 

Other module options seem to have a limited or mixed effect on structural bias in modCMA.

% \begin{itemize}
%     \item 435.456 mod CMA configurations (all module options)
%     \item BIAS evaluation using $10.000$ evaluations on $f_0$ in $5d$.
%     \item Deep-BIAS assessment due to time limitation of the original BIAS toolbox.
%     \item We focus on Centre SB, Bounds SB and Uniform (no SB), as cluster bias and discretization bias are often misclassified instances of centre bias and barely present in our data.
% \end{itemize}

\begin{figure*}[!tb]
\centering
 \includegraphics[width=0.243\textwidth,trim=15mm 20mm 20mm 20mm,clip]{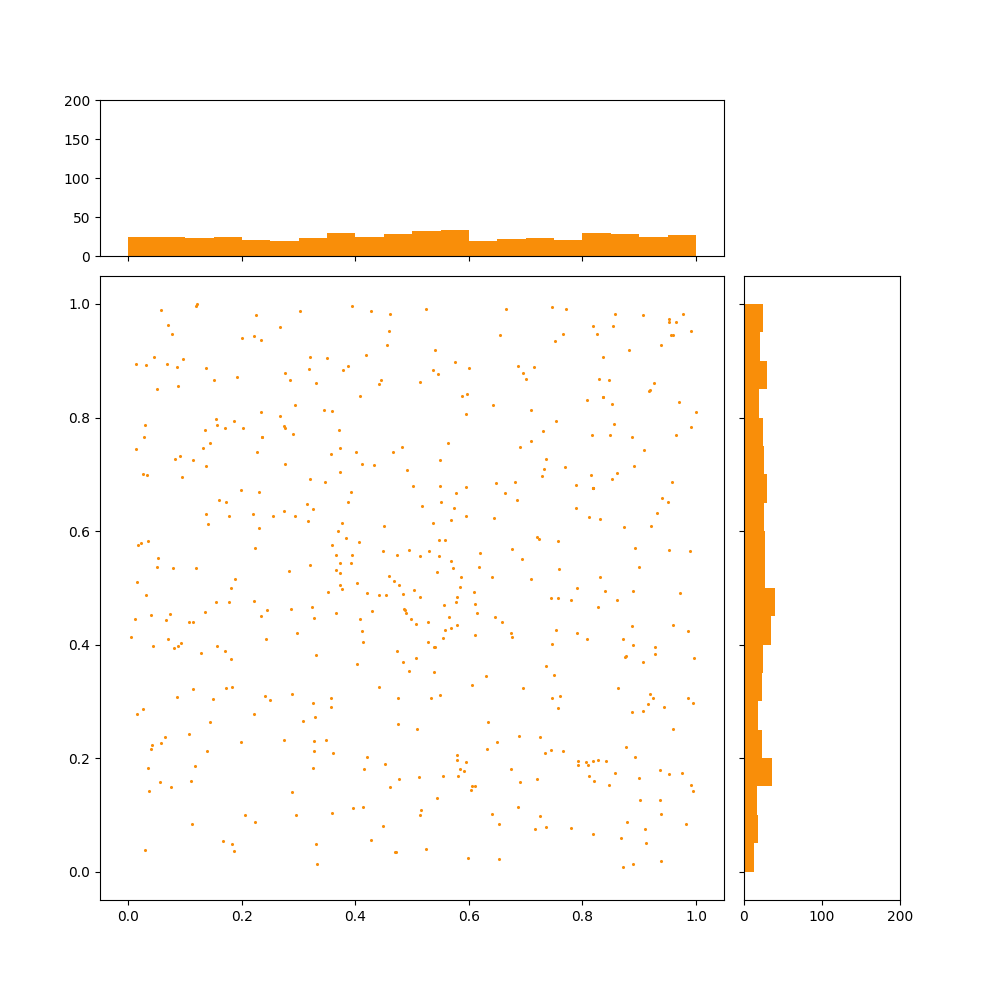}
 \includegraphics[width=0.243\textwidth,trim=15mm 20mm 20mm 20mm,clip]{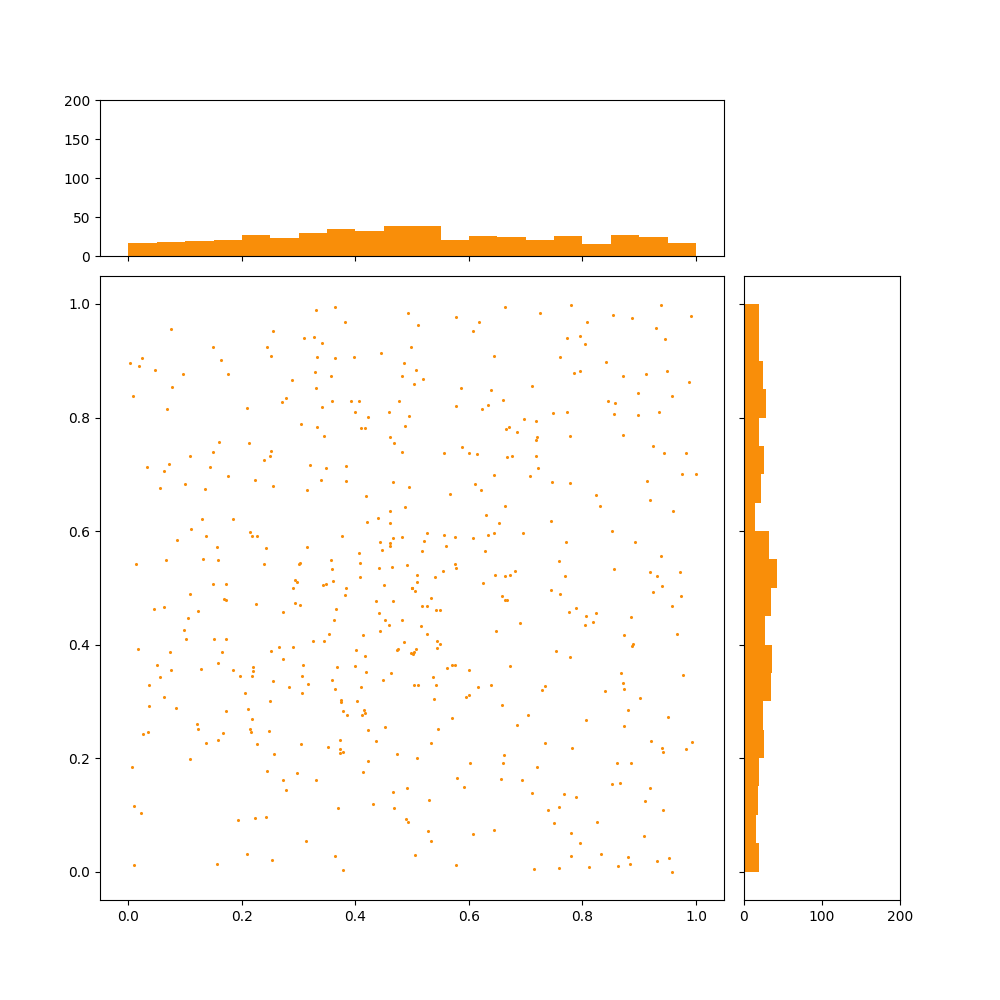}
 \includegraphics[width=0.243\textwidth,trim=15mm 20mm 20mm 20mm,clip]{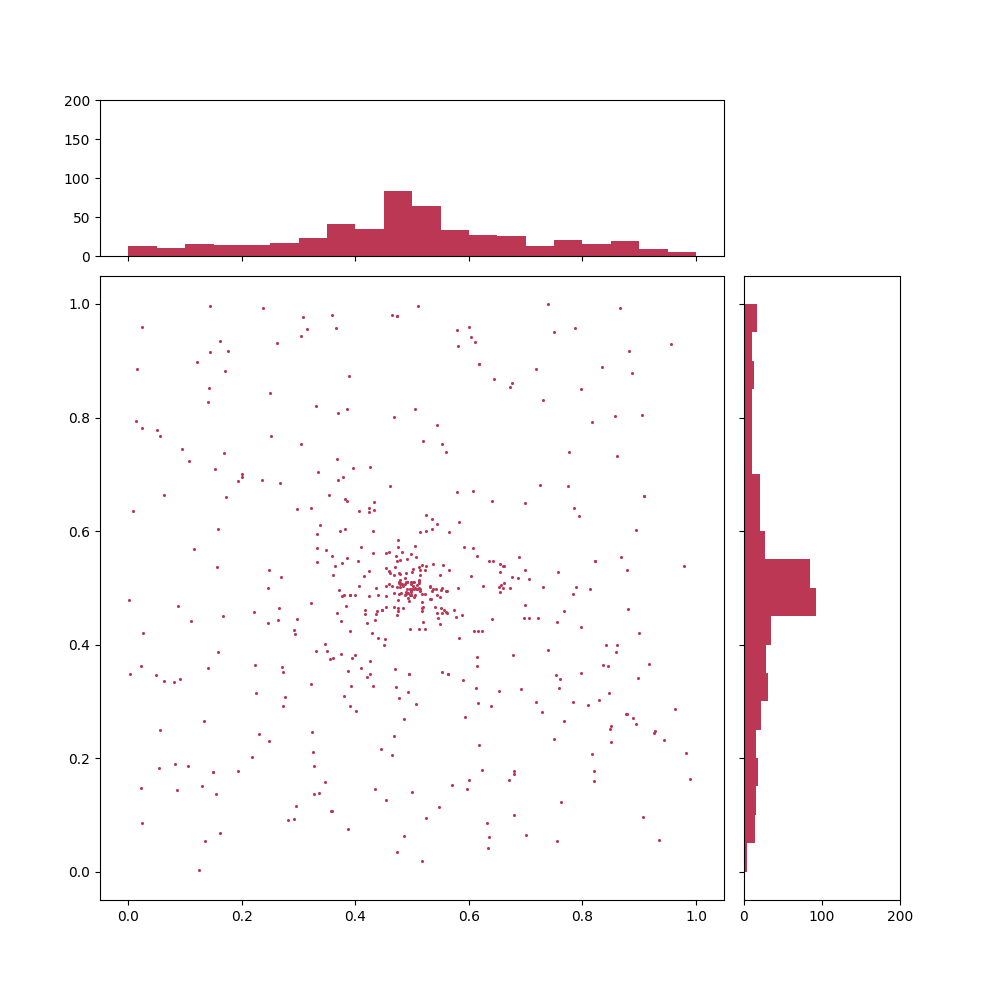}
 \includegraphics[width=0.243\textwidth,trim=15mm 20mm 20mm 20mm,clip]{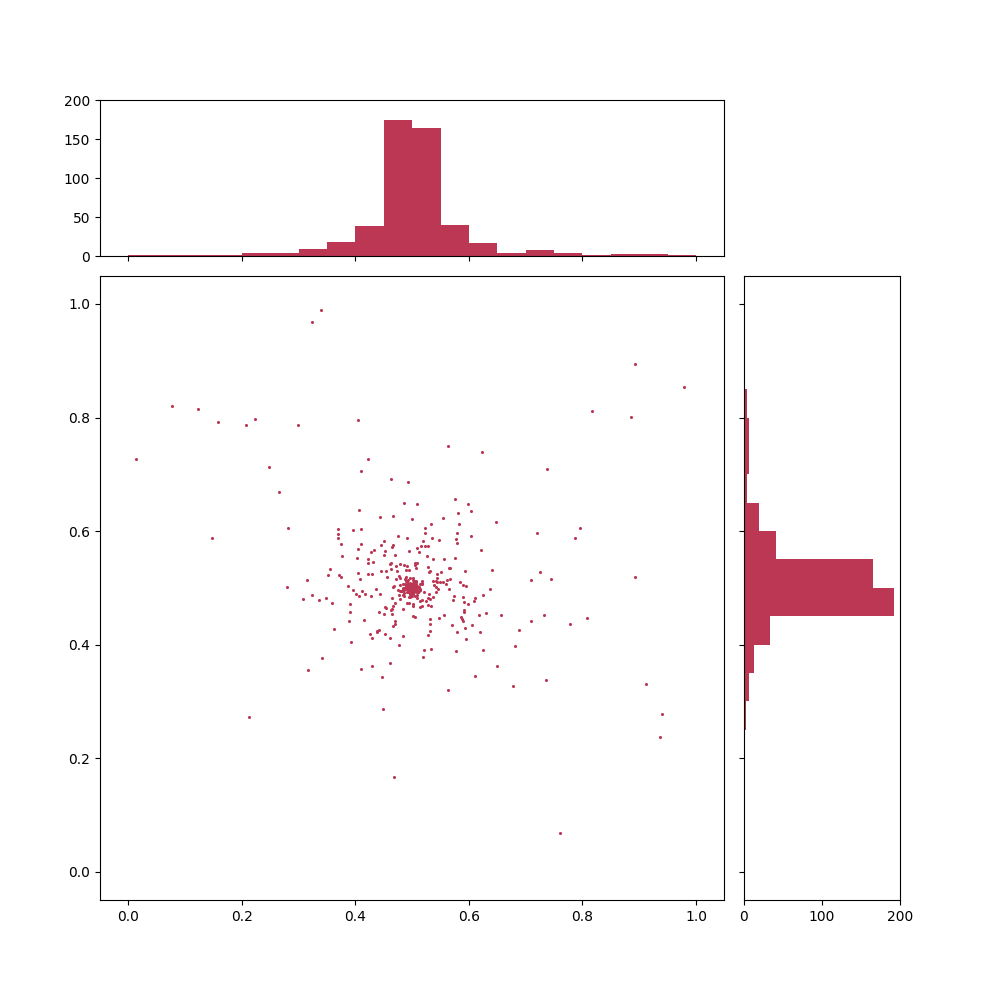}

 \includegraphics[width=0.243\textwidth,trim=15mm 20mm 20mm 20mm,clip]{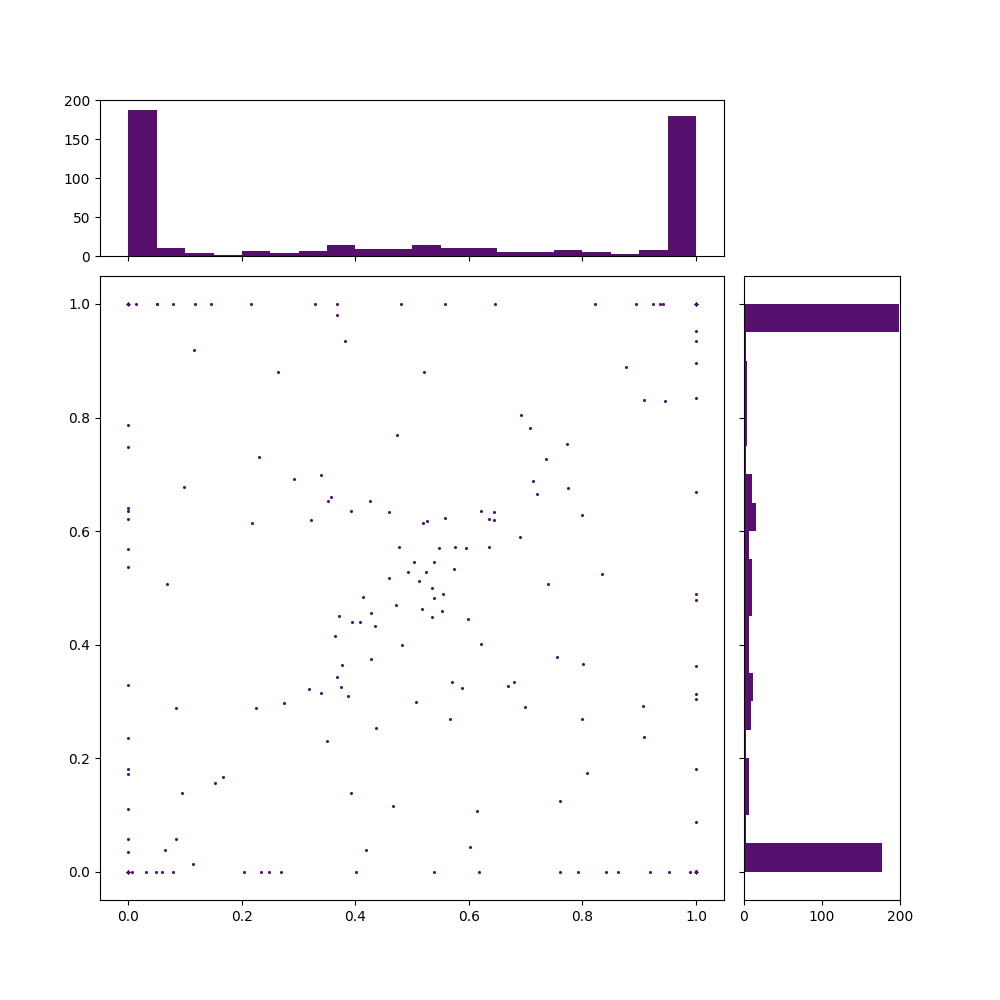}
 \includegraphics[width=0.243\textwidth,trim=15mm 20mm 20mm 20mm,clip]{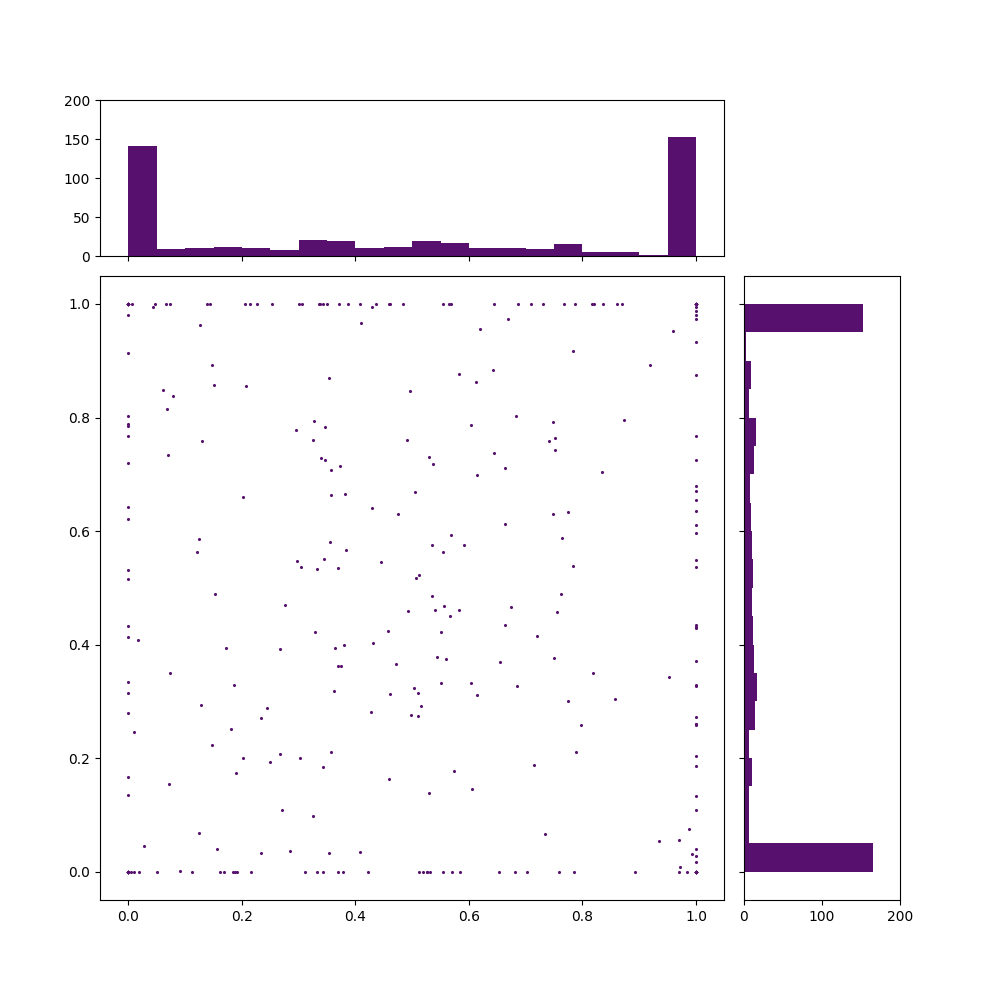}
 \includegraphics[width=0.243\textwidth,trim=15mm 20mm 20mm 20mm,clip]{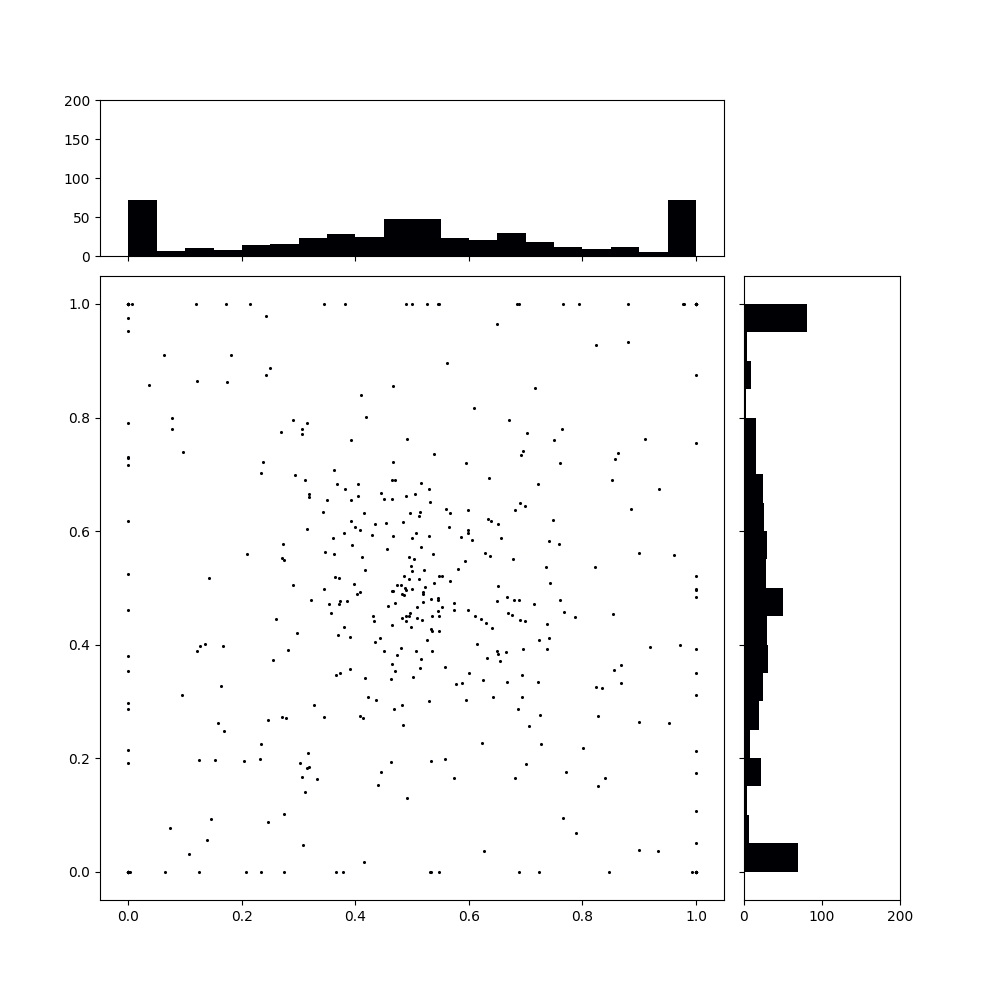}
  \includegraphics[width=0.243\textwidth,trim=15mm 20mm 20mm 20mm,clip]{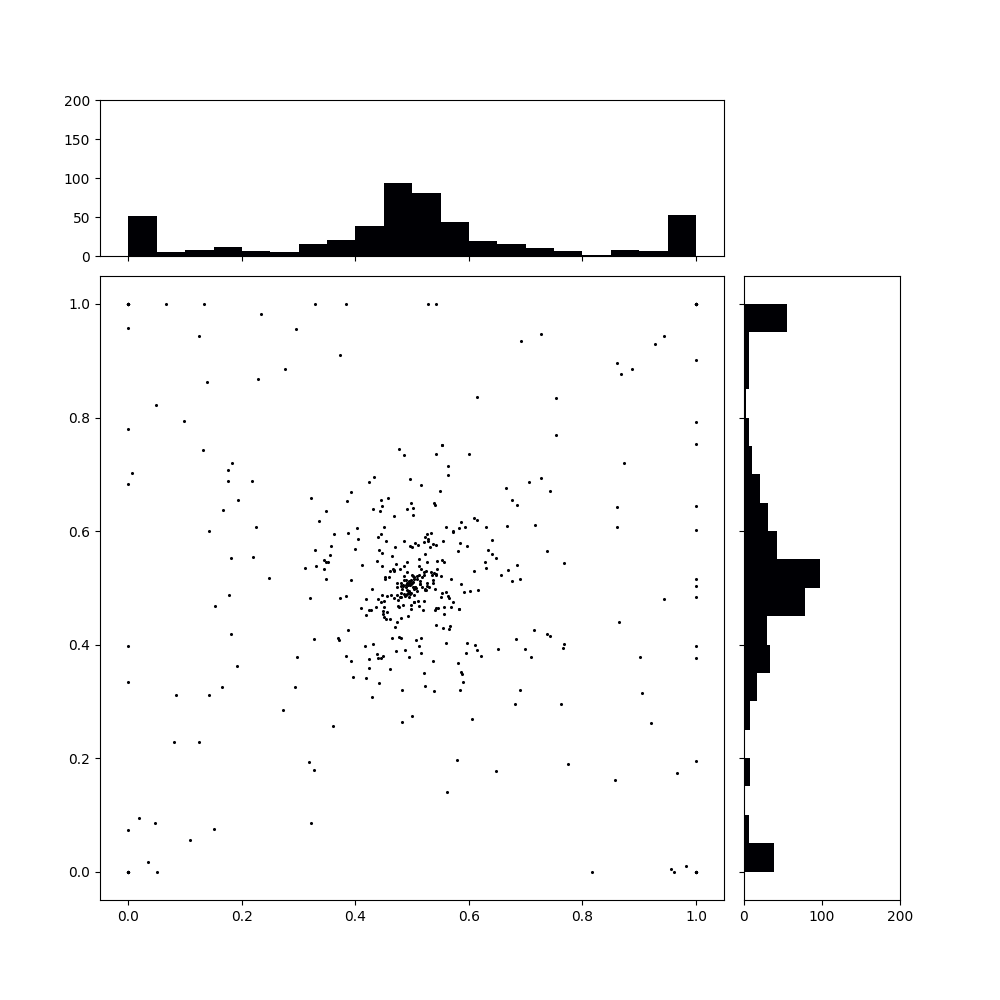}
\caption{Examples of the final best point distributions from CMA-ES configurations run on the SB test function $f_0$ in $2D$ from $500$ independent runs (using different random seed) of configurations (re)classified as \textbf{\textcolor[HTML]{f98e09}{no}}, \textbf{\textcolor[HTML]{bc3754}{centre}}, \textbf{\textcolor[HTML]{57106e}{bounds}} and \textbf{\textcolor[HTML]{000004}{mixed}} SB. \label{fig:distributions}}
\end{figure*}

\subsection{Limitations of Deep-BIAS and Mixed SB}
\label{sec:inspection}
While the deep-learning approach of the Deep-BIAS toolbox is very fast, and therefore allows the evaluation of $400\,000+$ configurations, it is not perfect. First of all, it is known \cite{van2023deep} that the SB type `Clusters' is often a misclassified `Centre' SB. As a result of this, after visual inspection of (a large fraction) of the configurations that were initially classified here as Cluster SB, we found out that all of those actually belonged to the Centre class. We therefore discarded the `Clusters' class in our analysis. In addition, after visual inspection of the configurations with highest confidence scores for each of the SB classes, we discovered a \emph{mixed} class between centre and bounds bias. This mix of two bias directions was not discovered in earlier structural bias research and was also not taken into account when developing the (Deep)-BIAS Toolbox. For modCMA, this mixed SB behaviour seems to occur when there is a configuration that is normally centre biased and it also uses the Saturate bound correction method (inducing additional bounds SB). For further analysis of the effects of these different SB classes on performance, we decided to re-classify the top $20$ configurations (sorted by confidence) for each SB class as identified by Deep-BIAS by visual inspection of the $2D$ final distributions into four SB classes: Uniform (no SB), Centre, Bounds and Mixed SB. See Figure \ref{fig:distributions} for examples of each class of SB we took into consideration. The complete set of configurations and distributions of their final best points across runs can be found in the supplemental material \cite{supplementalmaterial}.

\section{Effects of Structural Bias on Performance}

To analyse the effect of SB on algorithm performance, four distinct SB groups of algorithm configurations are evaluated on a range of affine function combinations. By gradually changing the function landscape properties, it is evaluated how the performance of these different groups changes under different conditions.

\subsection{Affine function pairs}
We include as original problems the \small{\textsc{sphere}} function (\emph{f1} from BBOB, uni-modal) and four other BBOB functions: \emph{f3} (separable \small{\textsc{Rastrign}}, multi-modal), \emph{f15} (non-seperable Rastrign, multi-modal), \emph{f16} (\small{\textsc{Weierstrass}}, multi-modal, adequate global structure), and \emph{f21} (Gallagher’s Gaussian 101-me \small{\textsc{Peaks}} Function, multi-modal, weak global structure). All of these are visualised in $2D$ in Figure 1 of the supplemental material. The sphere function was chosen because we would like to \emph{tune flatness into the affine combinations}, and the other four were chosen after visual inspection of their ruggedness (we would like to track structural bias along affine trajectories which begin at the flatness of the sphere function and \emph{gradually become more rugged}). 

We consider affine combinations between BBOB functions and use the generator proposed by Vermetten \emph{et al.} \cite{vermetten2023ma} which facilitates combinations of more than two functions --- although we consider only pairs here. Their generator takes three objects as input in order to construct a function: 1. the desired location for the optimum, $\overset{\scriptscriptstyle\rightarrow}{X}_{opt}$; 2. a vector of length 24  --- for each of the 24 BBOB functions --- indicating proportions, $\overset{\scriptscriptstyle\rightarrow}{W}$; and 3. a vector of length 24 indicating which instances of the BBOB functions should be used, $\overset{\scriptscriptstyle\rightarrow}{I}$. Of course, to obtain pairwise combinations then the proportions of 22 functions can be set to zero and the remaining two have non-zero weight. An affine combination \(\Xi\) is constructed by the generator according to the fitness \emph{scaling functions}:

\begin{equation}
 R_i(x) = \frac{max(\log_{10}(x),-8) + 8}{S_i}\\
\end{equation}

and its inverse (to reverse back to the original fitness scale):

\begin{equation}
R_i^{-1}(x) = 10^{\left(S_i \cdot x - 8\right)}
%R^{-1}(x) = 10^{(10.x)-8}
\end{equation}

$S_i$ is a scale factor and is set at literature-recommended values \cite{vermetten2023ma} depending on the base function: 11.0 for \emph{f1}, 12.3 for \emph{f3} and \emph{f15}, 10.3 for \emph{f16}, and 10.7 for \emph{f21}. 

With these defined, we can formally state that \(\Xi\) can be obtained by the generator as such:

\begin{equation}
\begin{split}
\Xi(\overset{\scriptscriptstyle\rightarrow}{W}, \overset{\scriptscriptstyle\rightarrow}{I}, \overset{\scriptscriptstyle\rightarrow}{X}_{opt}) = R^{-1}(\sum_{i=1}^{24}W_i.R_i(f_1,I_1(x - \overset{\scriptscriptstyle\rightarrow}{X}_{opt} + O_i,I_i) - f_i,I_i(O_i,I_i)))
\end{split}
\end{equation}

where $f_i,I_i$ is instance $i$ of original function $f_i$ and $O_i,I_i$ is the location of the optimum for instance $i$ of function $f_i$. \\

\subsection{Experimental setup}

We access the 24 noiseless BBOB functions in $2D$ through \textsc{IOHexperimenter} \cite{de2024iohexperimenter}. 

\subsubsection{Affine combinations}
The original BBOB functions involved in the affine pairs are all $2D$, to facilitate visualisation. We consider the region of interest \([-4, 4]\) per dimension only.
For each function pair, a sequence of 51 values $\alpha \in[0,1]$ is defined equally spaced with a step of 0.02. We define \(\alpha\) as the \emph{proportion of the \small{\textsc{Sphere}} function}, which is BBOB \emph{f1}. 
For each combination of two functions with a given alpha, we generate four affine combinations which differ only in the location of the global optimum --- we use the same instances of the BBOB constituent parts for each of them. We also keep the instance number and optimum location consistent across increasing \(\alpha\) within each combination of base function pair and optimum placement strategy. Instances are randomly generated between 1 and 100. The four placement strategies for the optimum are:

\begin{enumerate}
    \item near to the boundary (within 0.01) in both coordinates,
    \item near the centre (between \([-0.01, 0.01]\) in both coordinates),
    \item near to the boundary in one coordinate and central in the other,
    \item located randomly for both coordinates between \([-2, 2]\).
\end{enumerate}

In total, we generate 816 affine recombination functions (4 original function pairs $\times$ 51 affine weights $\times$ 4 locations for the optimum). The process of affine combination and placement of the optimum is conducted using functions from the \textsc{IOHExperimenter} package in Python. 

\subsubsection{Algorithm performance}

For assessing algorithm performance on the 816 functions we consider the top-scoring modCMA configurations for each bias type after careful visual inspection of the SB distributions (See Section \ref{sec:inspection}) (\textbf{bounds} (20 configurations), \textbf{centre} (19), \textbf{mixed} (7), and \textbf{none} (10)). In total, this amounts to 56 CMAES variants and \num{45696} algorithm and function-pairs. The algorithm configurations for these are available in Tables 1-4 of the supplemental material \cite{supplementalmaterial}. Each CMA-ES configuration is instantiated in \textsc{modCMA}, provided a budget of \num{5000} evaluations, and is executed 30 times on each of the 816 affine functions. As the performance metric, we use a normalised area under the curve (AUC) with respect to the empirical cumulative distribution function, implemented by Vermetten \emph{et al}\footnote{\url{https://zenodo.org/records/10376912}}. The distribution function considers the default COCO settings with 51 targets spaced logarithmically beginning at \(10^{-8}\) and terminating at \(10^2\).

\begin{figure}[!t]
\centering
\includegraphics[width=\textwidth,trim=12mm 10mm 10mm 8mm,clip]{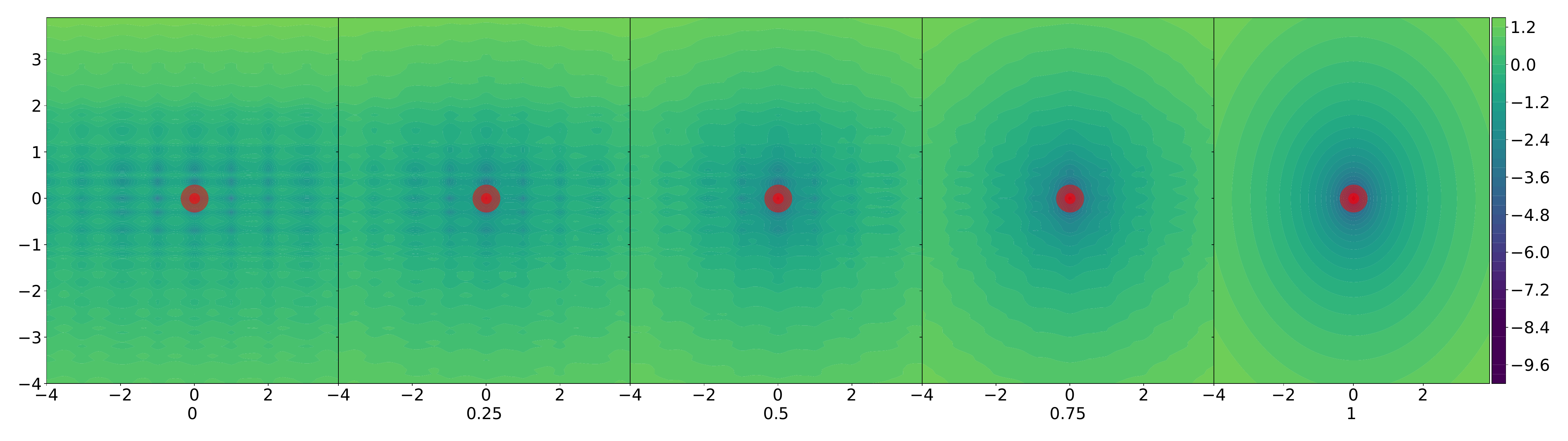}\label{fig:sphere-srast-affine}
\includegraphics[width=\textwidth,trim=12mm 10mm 10mm 8mm,clip]{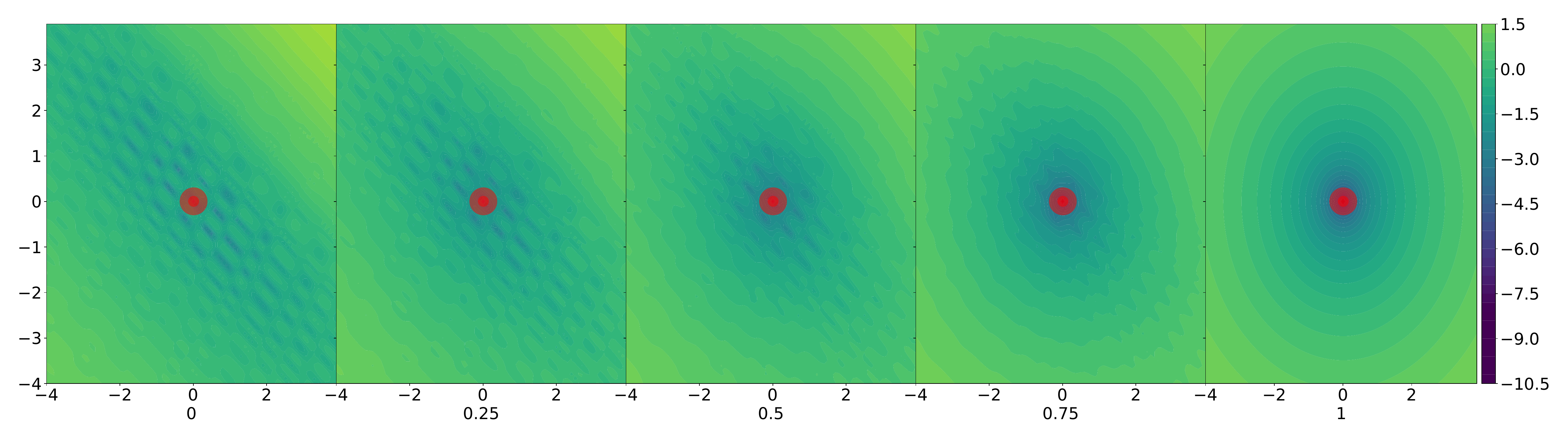}\label{fig:sphere-nrast-affine}
\includegraphics[width=\textwidth,trim=12mm 10mm 10mm 8mm,clip]{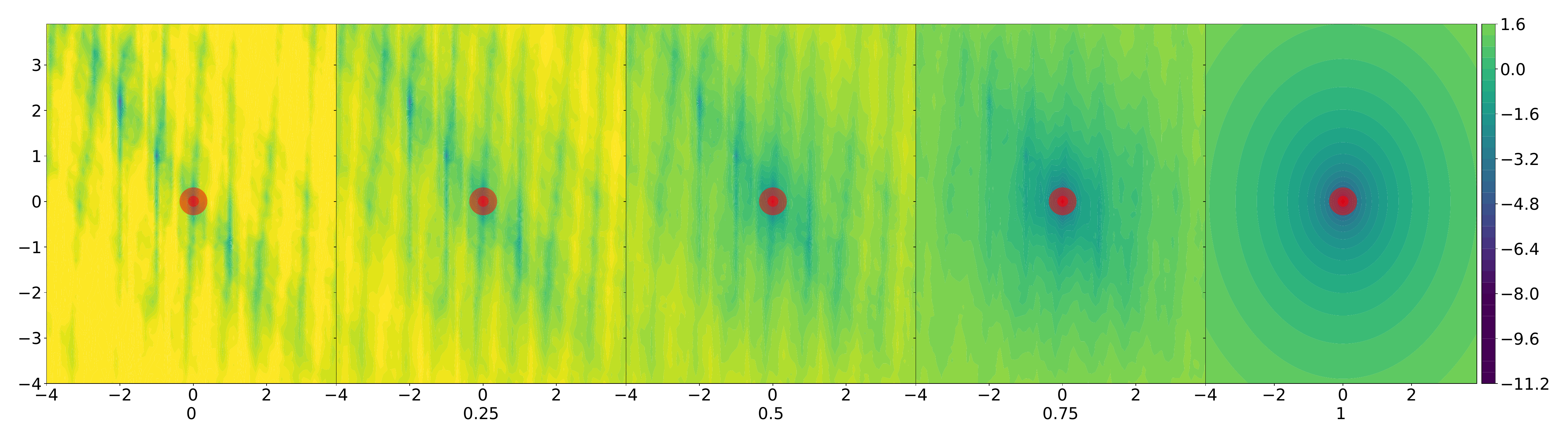}\label{fig:sphere-weier-affine}
\includegraphics[width=\textwidth,trim=12mm 10mm 10mm 8mm,clip]{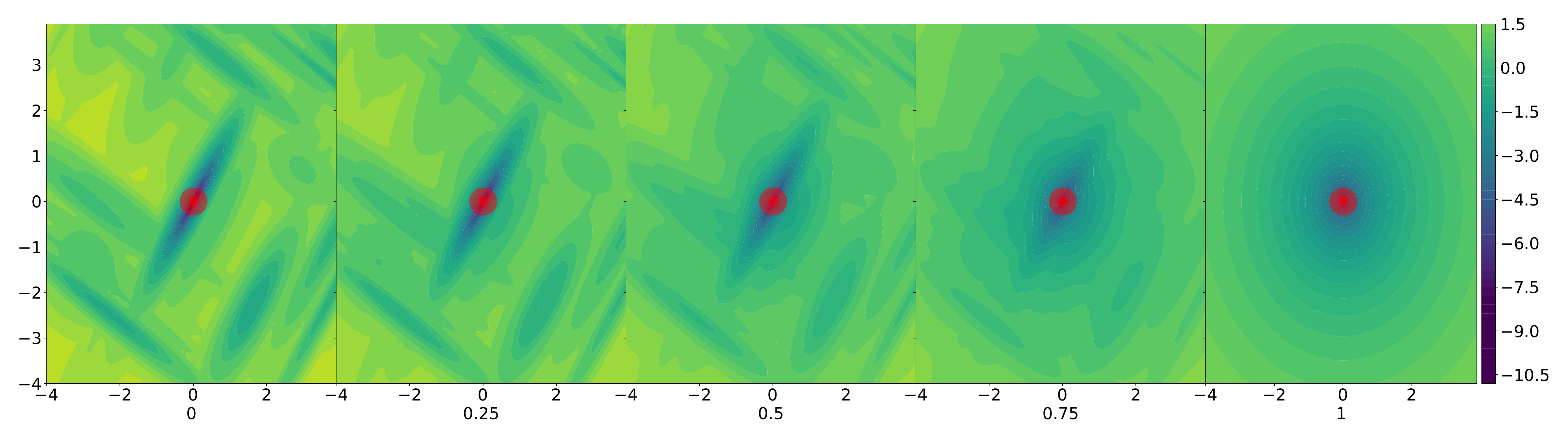}\label{fig:sphere-gall-affine}
\caption{Example $2D$ landscapes for affine combinations of functions \emph{f3}, \emph{f15}, \emph{f16}, \emph{f21} (top to bottom) with \emph{f1} for 5 affine weights $\alpha$ shown as labels below individual plots, where increasing \(\alpha\) corresponds to increasing the proportion of \emph{f1}. On the instances shown here, the location of the global optimum is fixed near the centre of the domain and marked in red. \label{fig:function-viz}} %From left to right, there is increasing \(\alpha\), which is the proportion of \emph{f1}. The location of the global optimum is shown at red; it has been placed near centre of the function 
\end{figure}

\begin{figure}[!t]
\centering
    \begin{subfigure}[b]{0.255\textwidth}
        \includegraphics[height=2.5cm,trim=15mm 25mm 30mm 24mm,clip]{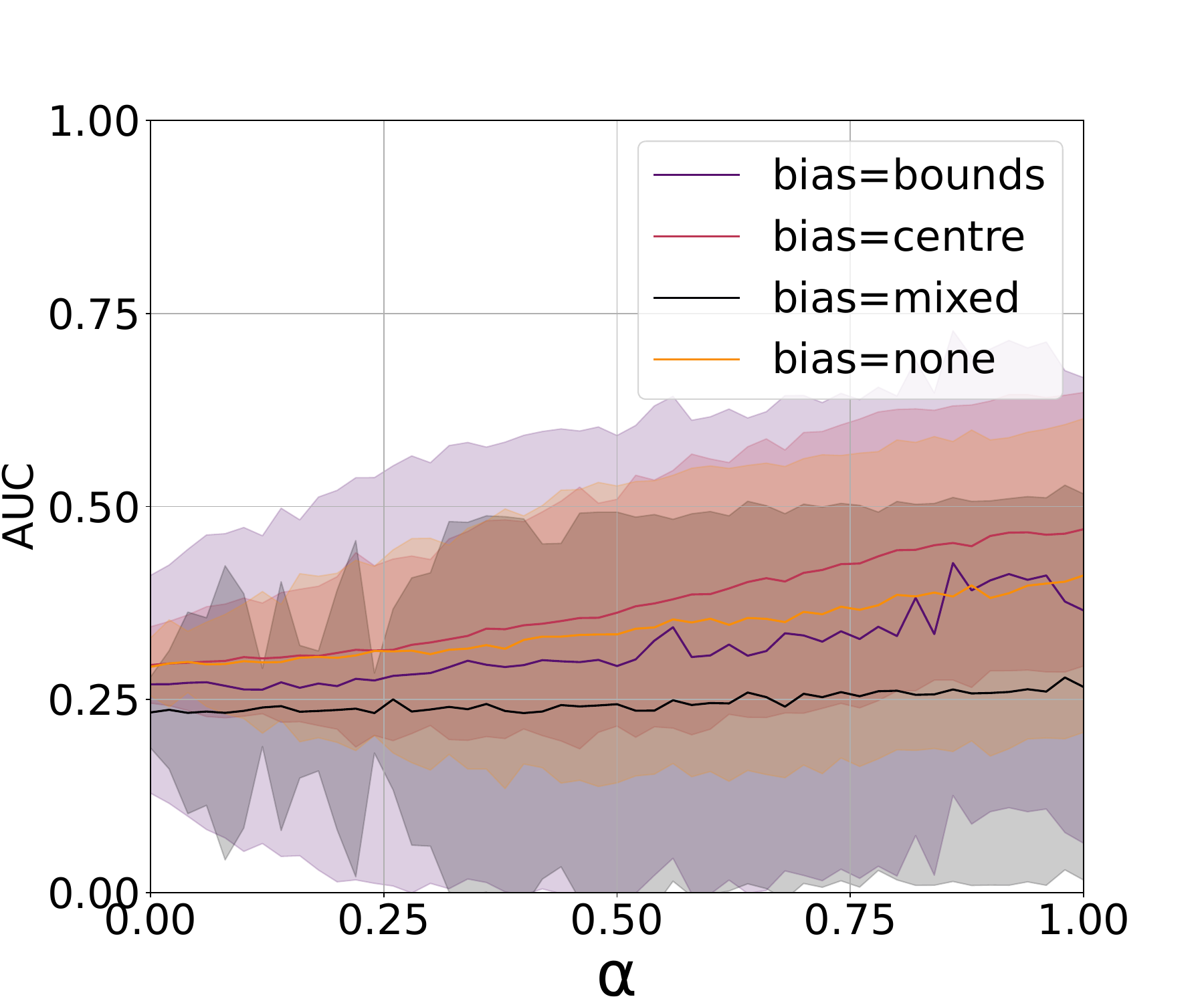}
        %\caption{Bounds}
        \label{fig:sa-1}
    \end{subfigure}
    \hfill
    \begin{subfigure}[b]{0.24\textwidth}
         \includegraphics[height=2.5cm,trim=35mm 25mm 30mm 24mm,clip]{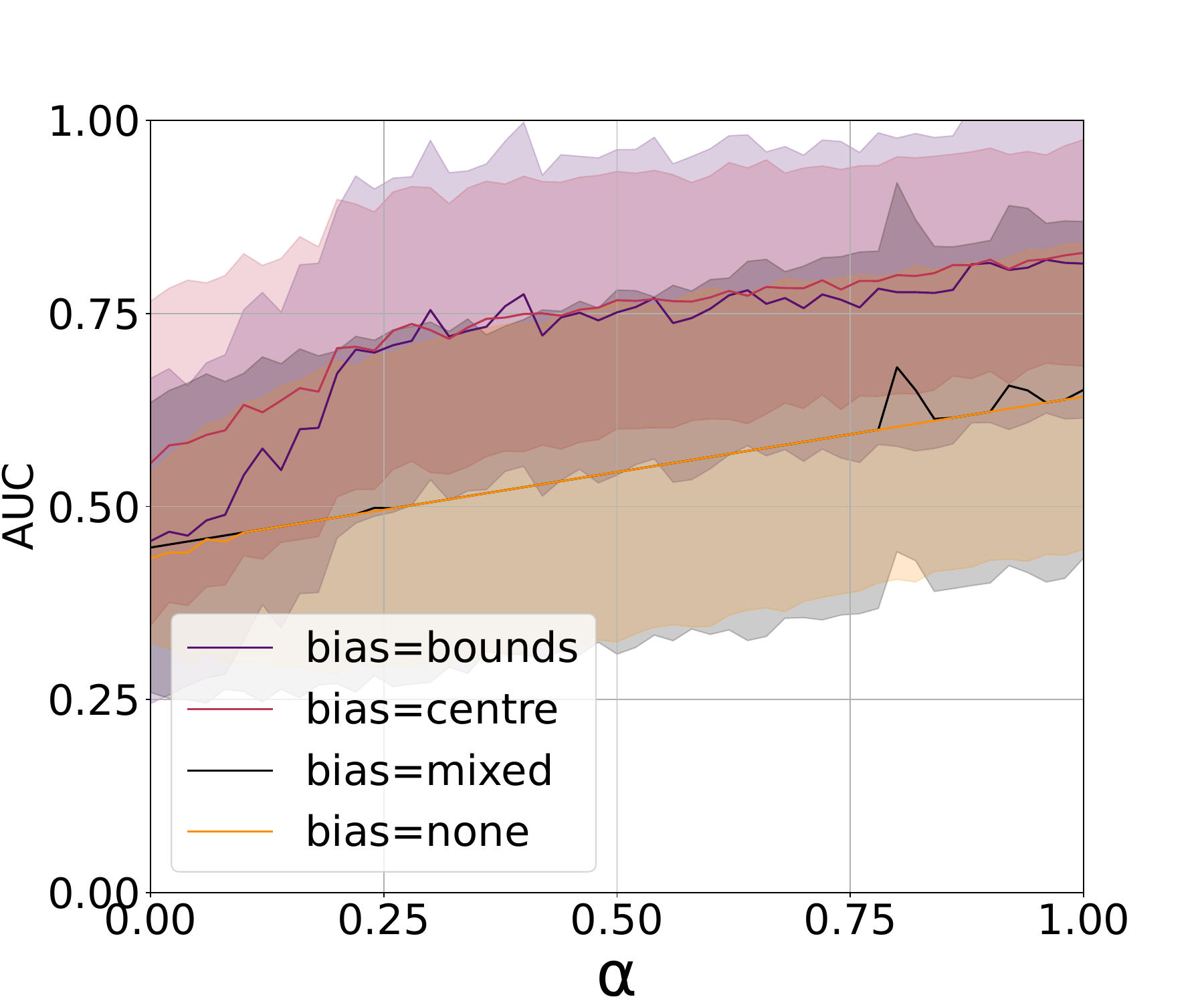}
        %\caption{Centre}
        \label{fig:sa-2}
    \end{subfigure}
    \begin{subfigure}[b]{0.24\textwidth}
       \includegraphics[height=2.5cm,trim=35mm 25mm 30mm 24mm,clip]{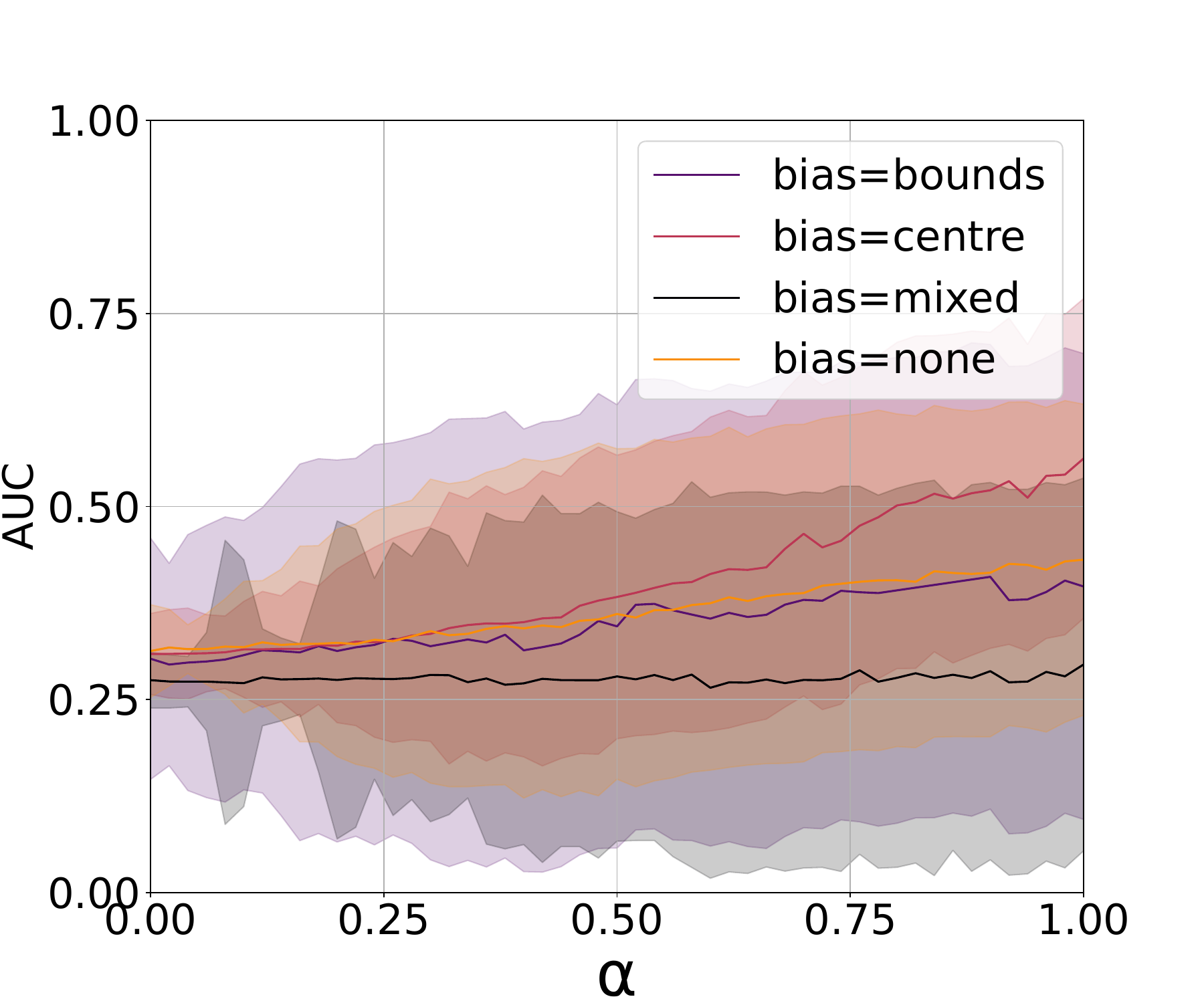}
        %\caption{Centre of bounds}
        \label{fig:sa-3}
    \end{subfigure}
    \begin{subfigure}[b]{0.24\textwidth}
       \includegraphics[height=2.5cm,trim=35mm 25mm 30mm 24mm,clip]{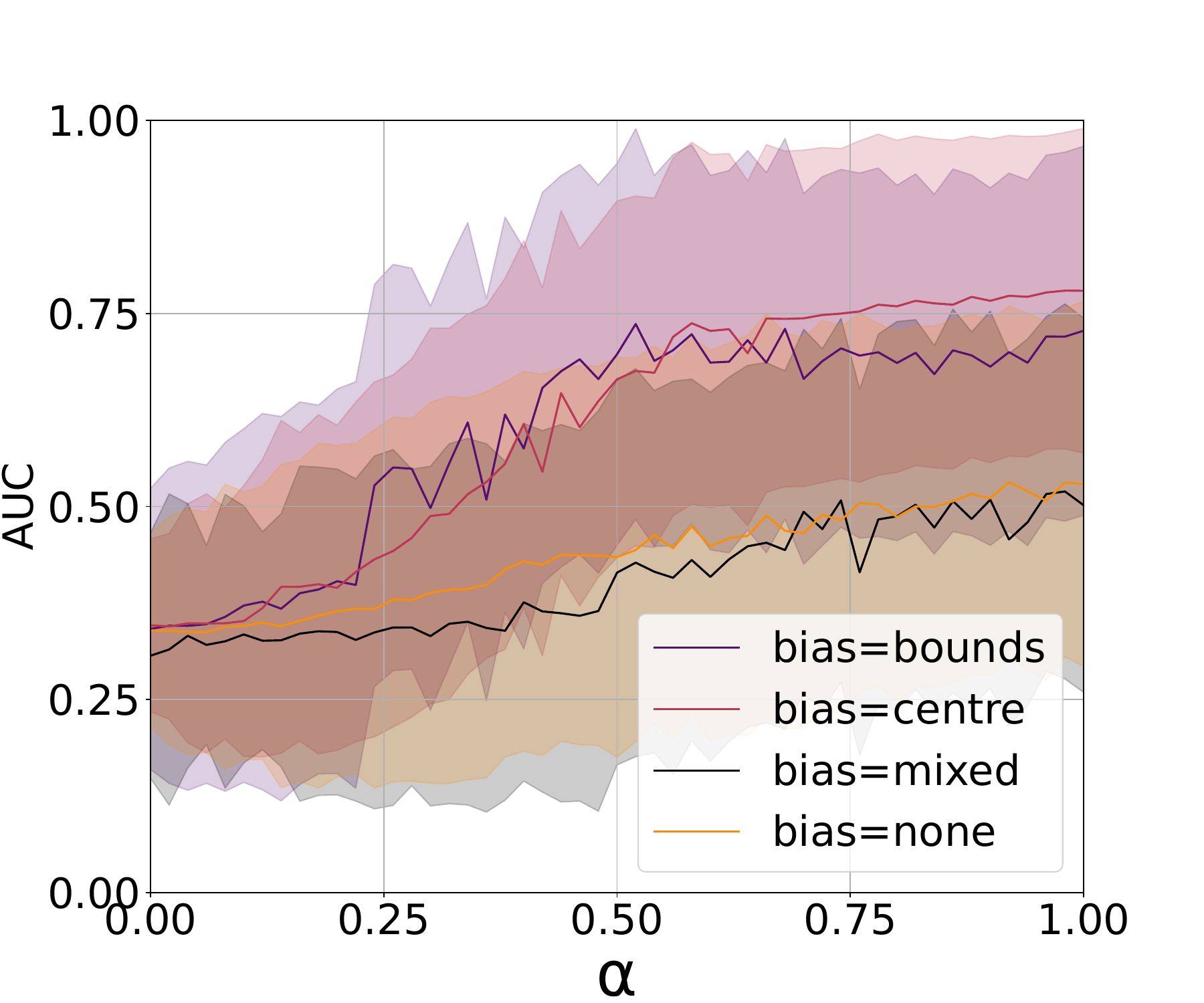}
        %\caption{Random}
        \label{fig:sa-4}
    \end{subfigure}
    \begin{subfigure}[b]{0.255\textwidth}
        \includegraphics[height=2.5cm,trim=15mm 25mm 30mm 24mm,clip]{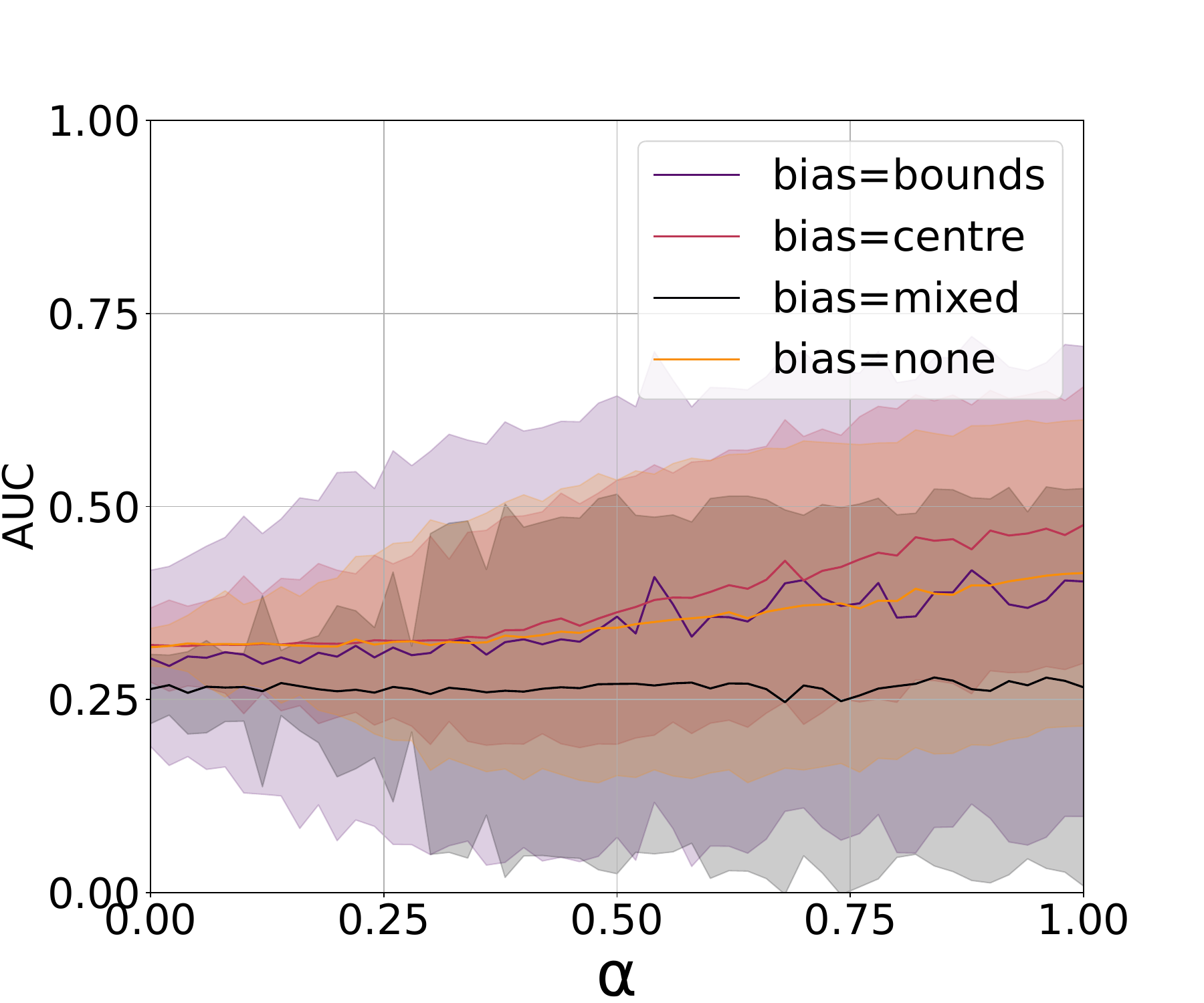}
        %\caption{Bounds}
        \label{fig:nrast-1}
    \end{subfigure}
    %\hfill % optional: you might not need this if the total width is exactly \textwidth
    \begin{subfigure}[b]{0.24\textwidth}
      \includegraphics[height=2.5cm,trim=35mm 25mm 30mm 24mm,clip]{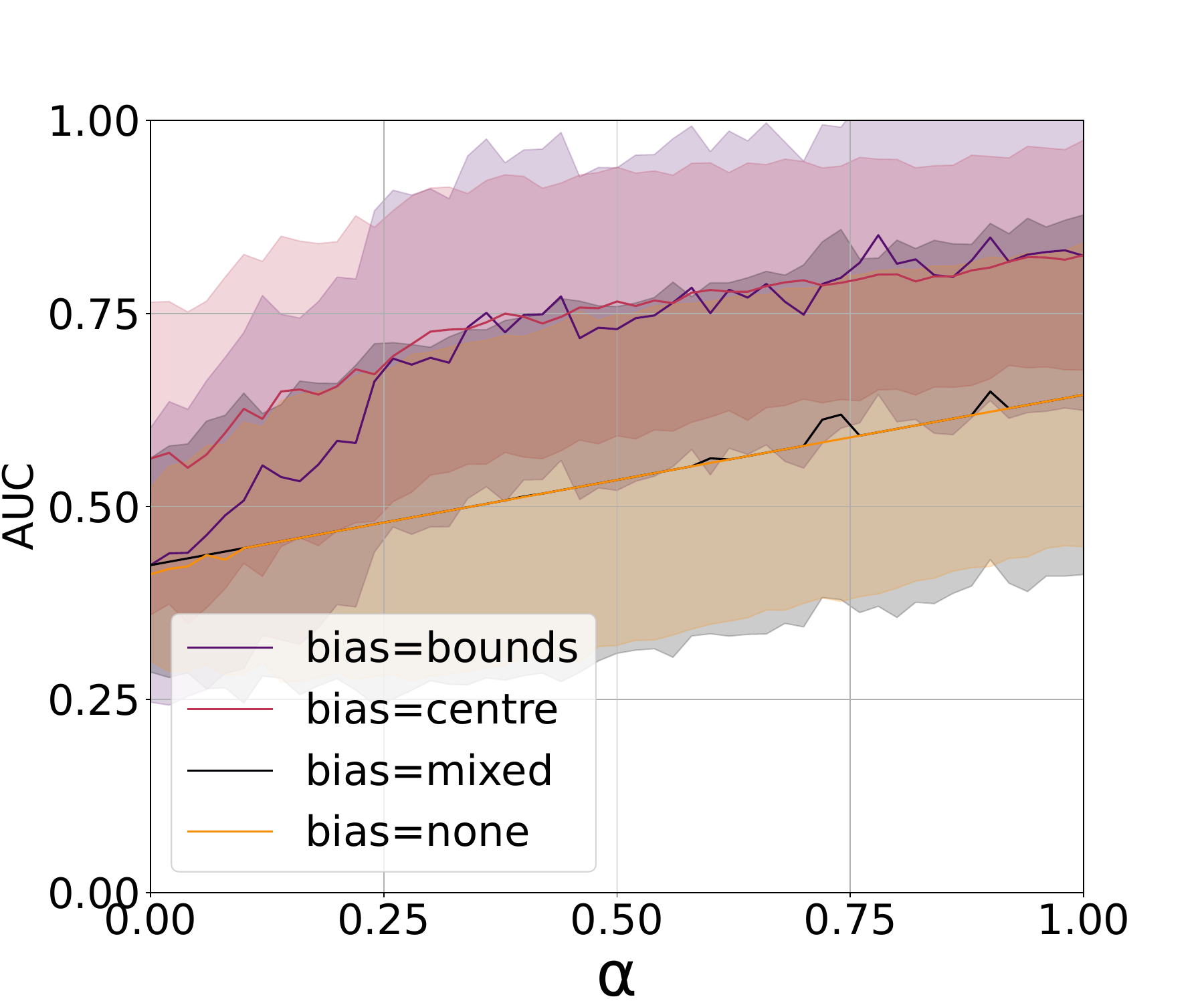}
        %\caption{Centre}
        \label{fig:nrast-2}
    \end{subfigure}
    \begin{subfigure}[b]{0.24\textwidth}
         \includegraphics[height=2.5cm,trim=35mm 25mm 30mm 24mm,clip]{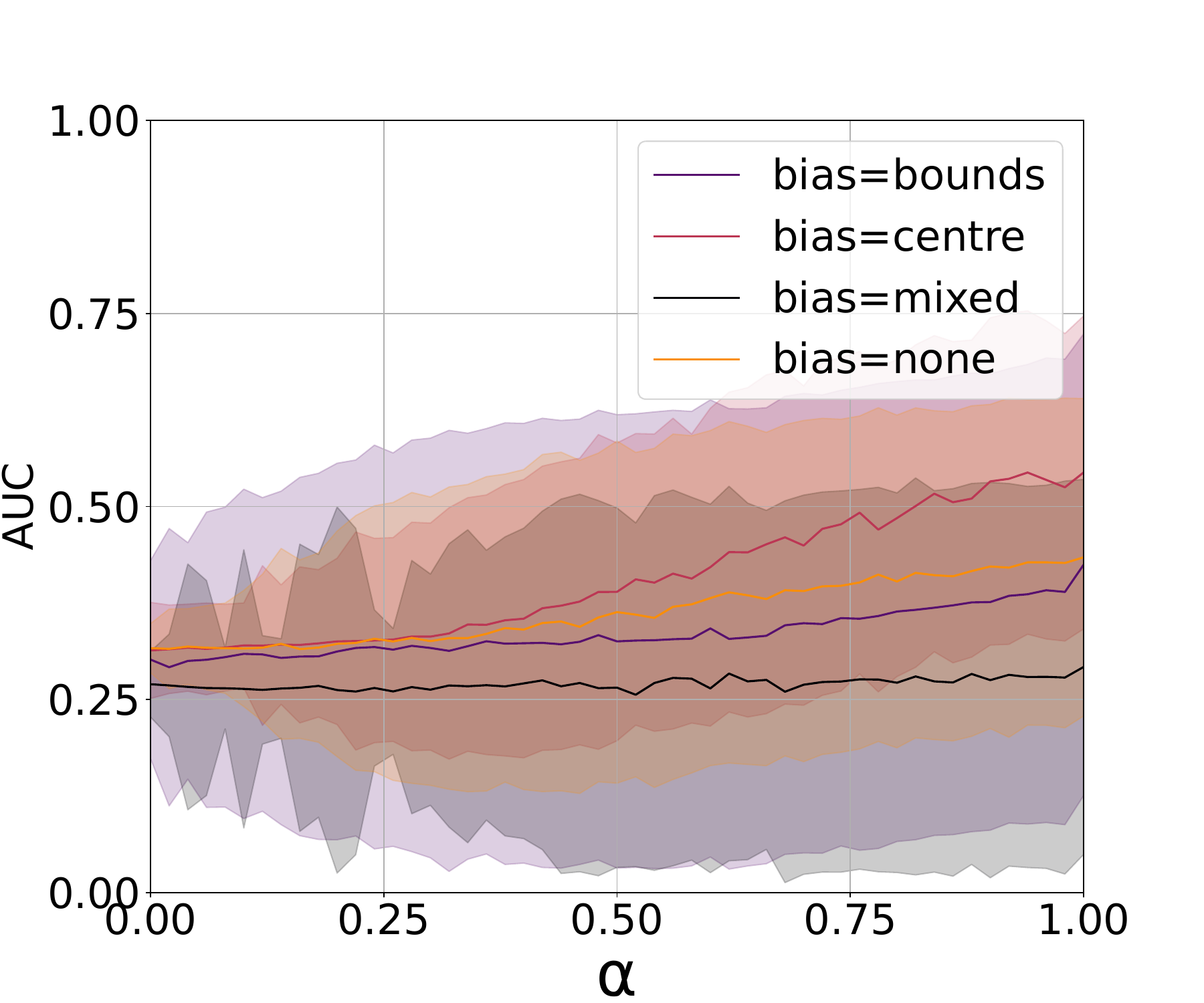}
        %\caption{Centre of bounds}
        \label{fig:nrast-3}
    \end{subfigure}
    %\hfill
    \begin{subfigure}[b]{0.24\textwidth}
         \includegraphics[height=2.5cm,trim=35mm 25mm 30mm 24mm,clip]{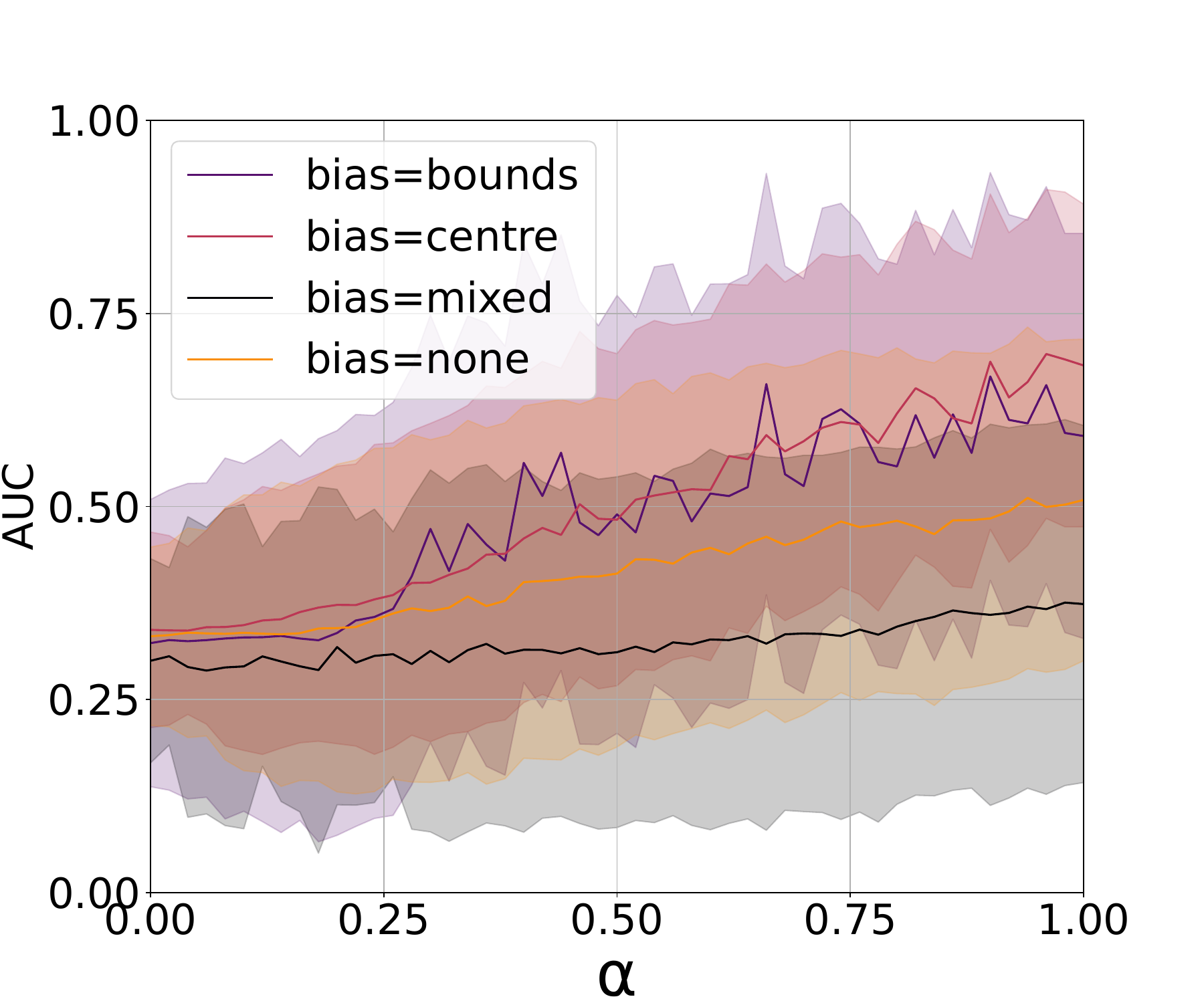}
        %\caption{Random}
        \label{fig:nrast-4}
    \end{subfigure}
    \begin{subfigure}[b]{0.255\textwidth}
         \includegraphics[height=2.5cm,trim=15mm 25mm 30mm 24mm,clip]{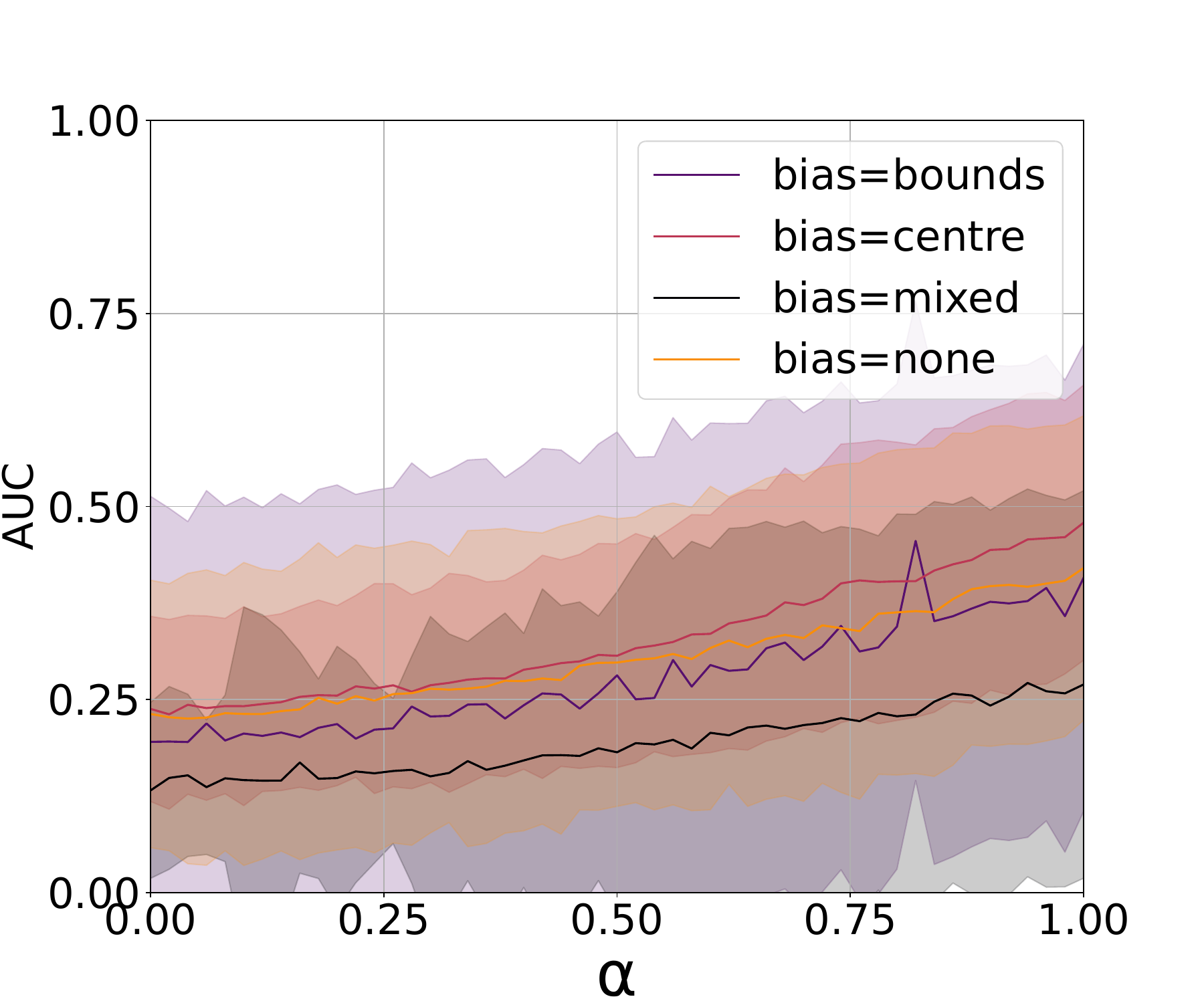}
        %\caption{Bounds}
        \label{fig:weier-1}
    \end{subfigure}
    %\hfill % optional: you might not need this if the total width is exactly \textwidth
    \begin{subfigure}[b]{0.24\textwidth}
        \includegraphics[height=2.5cm,trim=35mm 25mm 30mm 24mm,clip]{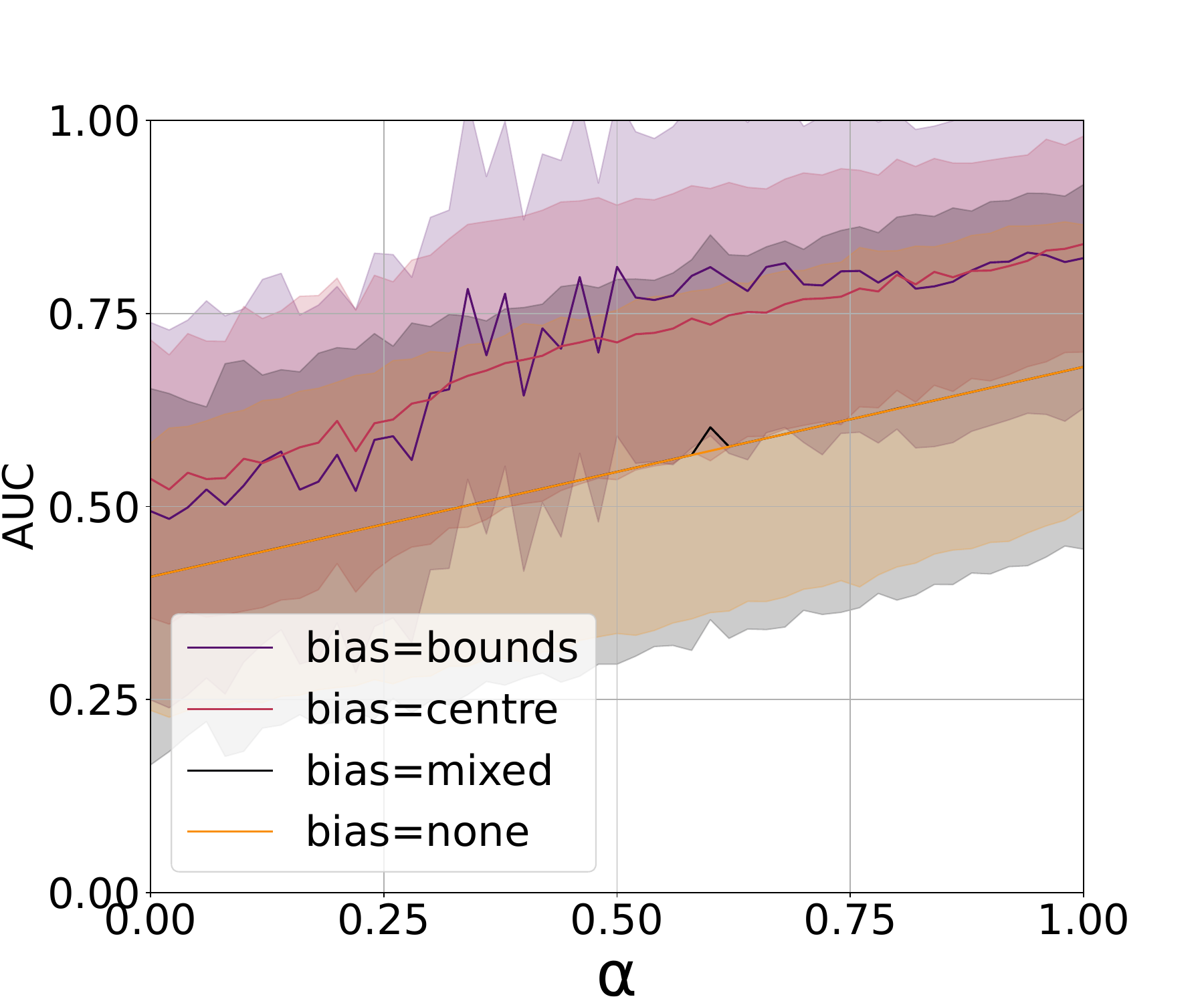}
        %\caption{Centre}
        \label{fig:weier-2}
    \end{subfigure}
    \begin{subfigure}[b]{0.24\textwidth}
         \includegraphics[height=2.5cm,trim=35mm 25mm 30mm 24mm,clip]{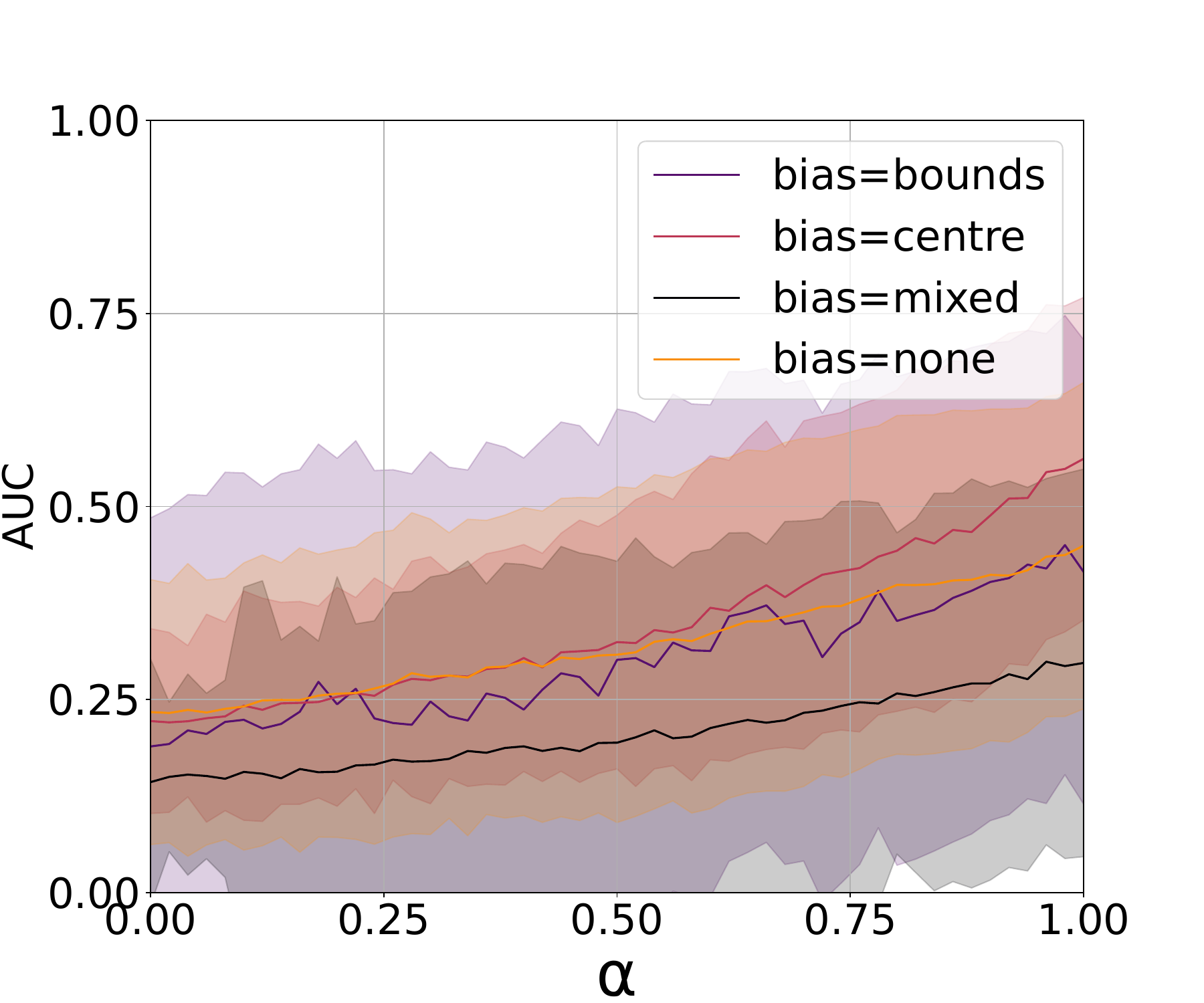}
        %\caption{Centre of bounds}
        \label{fig:weier-3}
    \end{subfigure}
    \begin{subfigure}[b]{0.24\textwidth}
         \includegraphics[height=2.5cm,trim=35mm 25mm 30mm 24mm,clip]{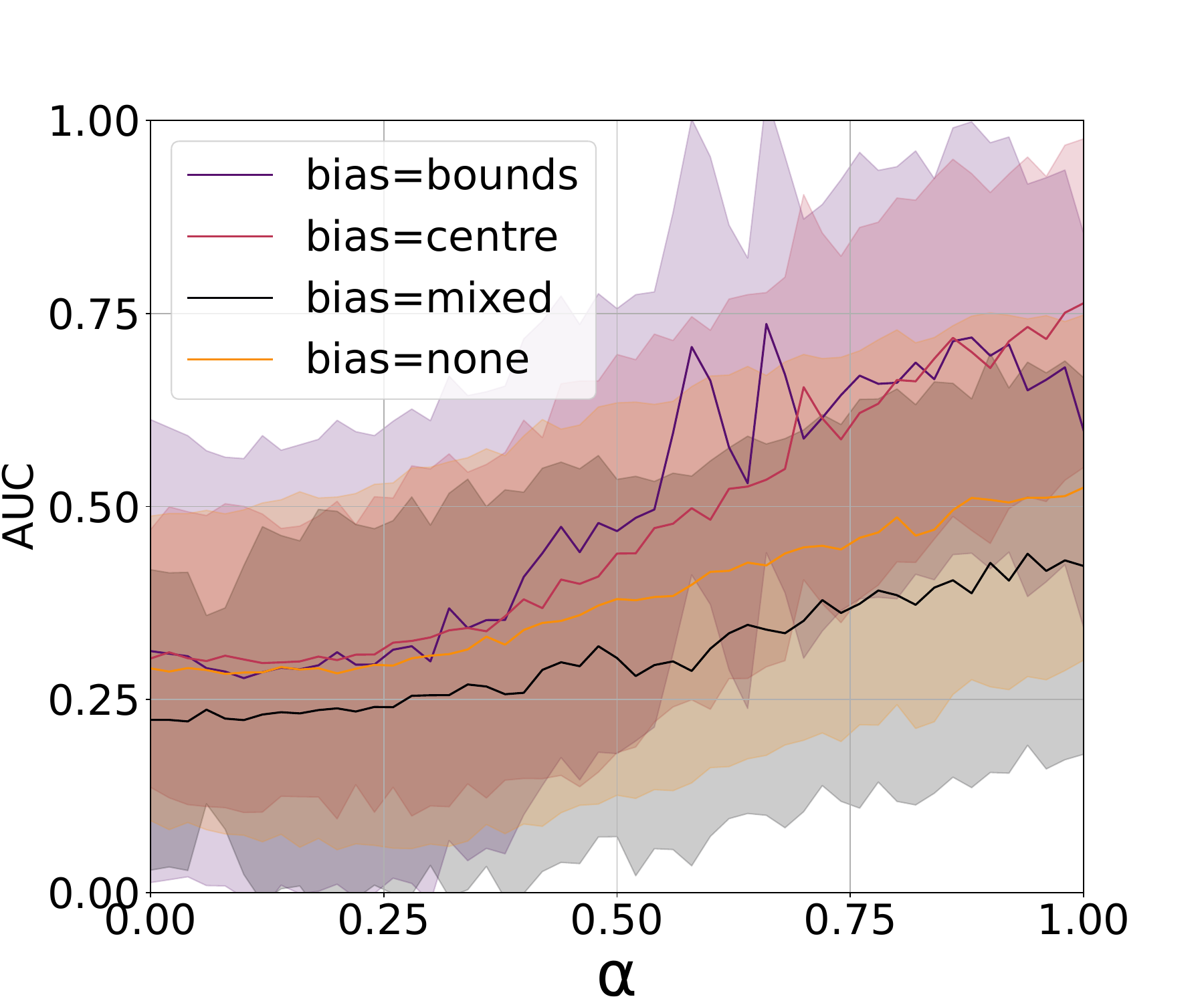}
        %\caption{Random}
        \label{fig:weier-4}
    \end{subfigure}
    \begin{subfigure}[b]{0.255\textwidth}
         \includegraphics[height=2.6cm,trim=15mm 15mm 30mm 24mm,clip]{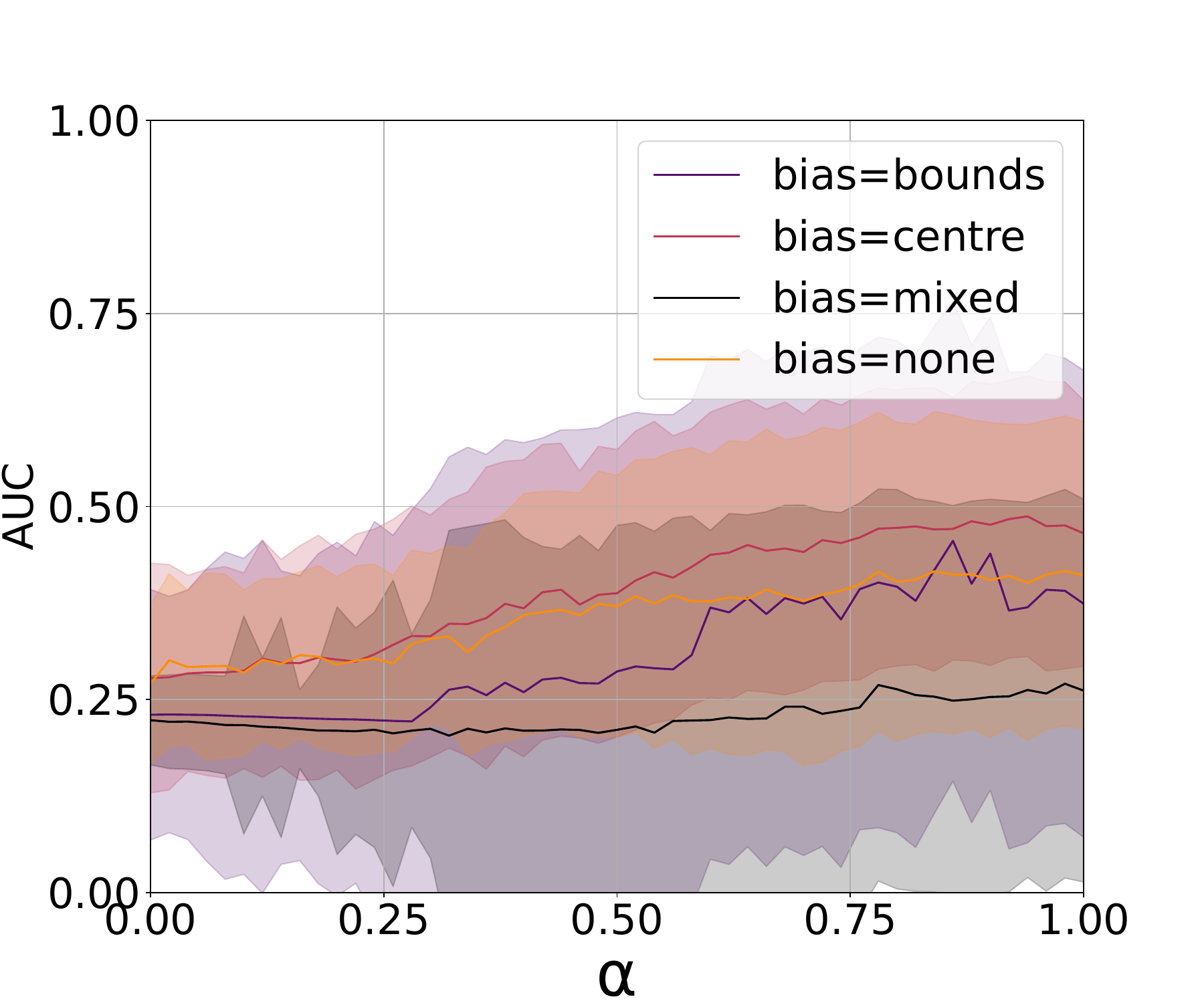}
        %\caption{Bounds}
        \label{fig:gall-1}
    \end{subfigure}
    \begin{subfigure}[b]{0.24\textwidth}
        \includegraphics[height=2.6cm,trim=35mm 15mm 30mm 24mm,clip]{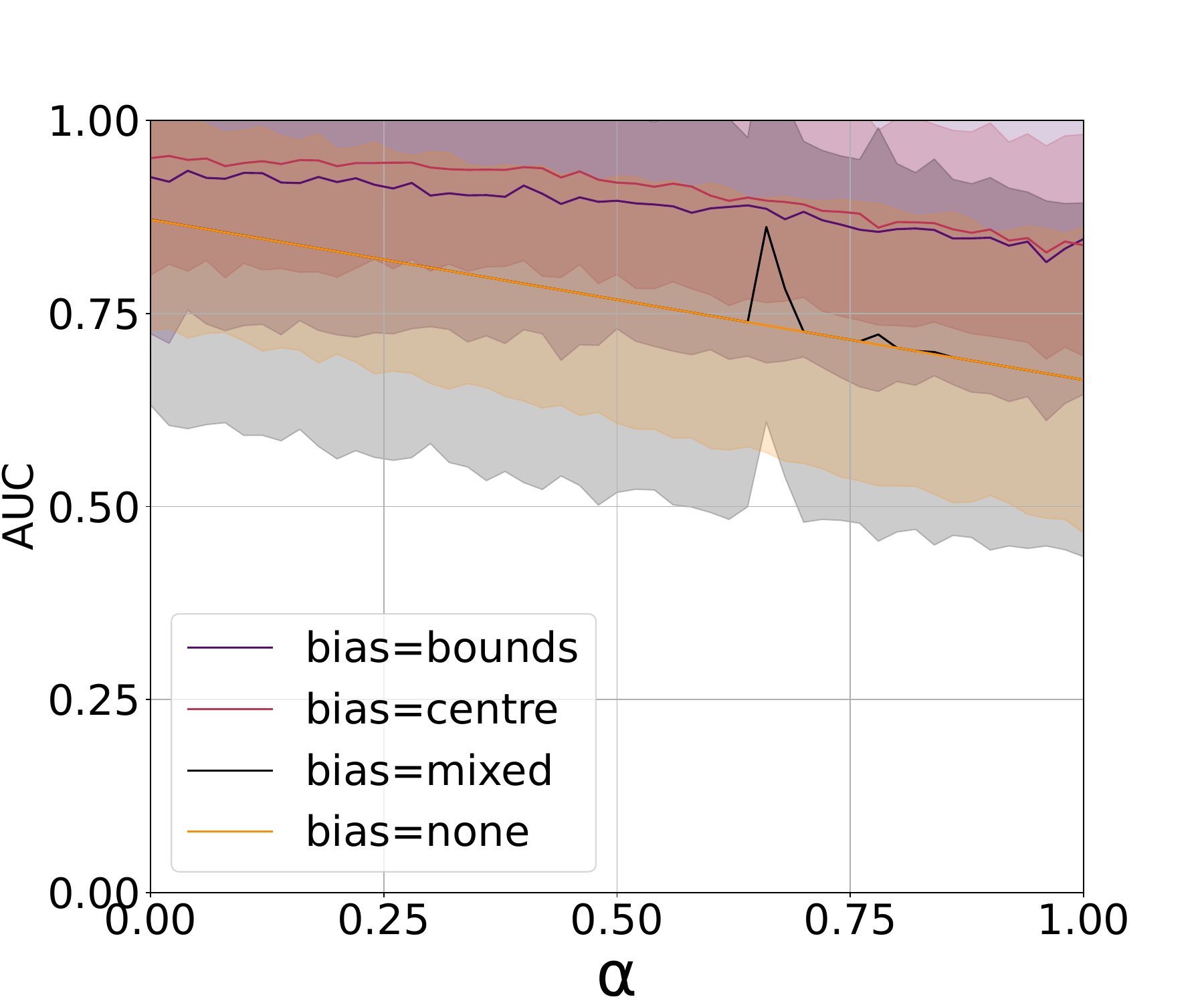}
        %\caption{Centre}
        \label{fig:gall-2}
    \end{subfigure}
    \begin{subfigure}[b]{0.24\textwidth}
         \includegraphics[height=2.6cm,trim=35mm 15mm 30mm 24mm,clip]{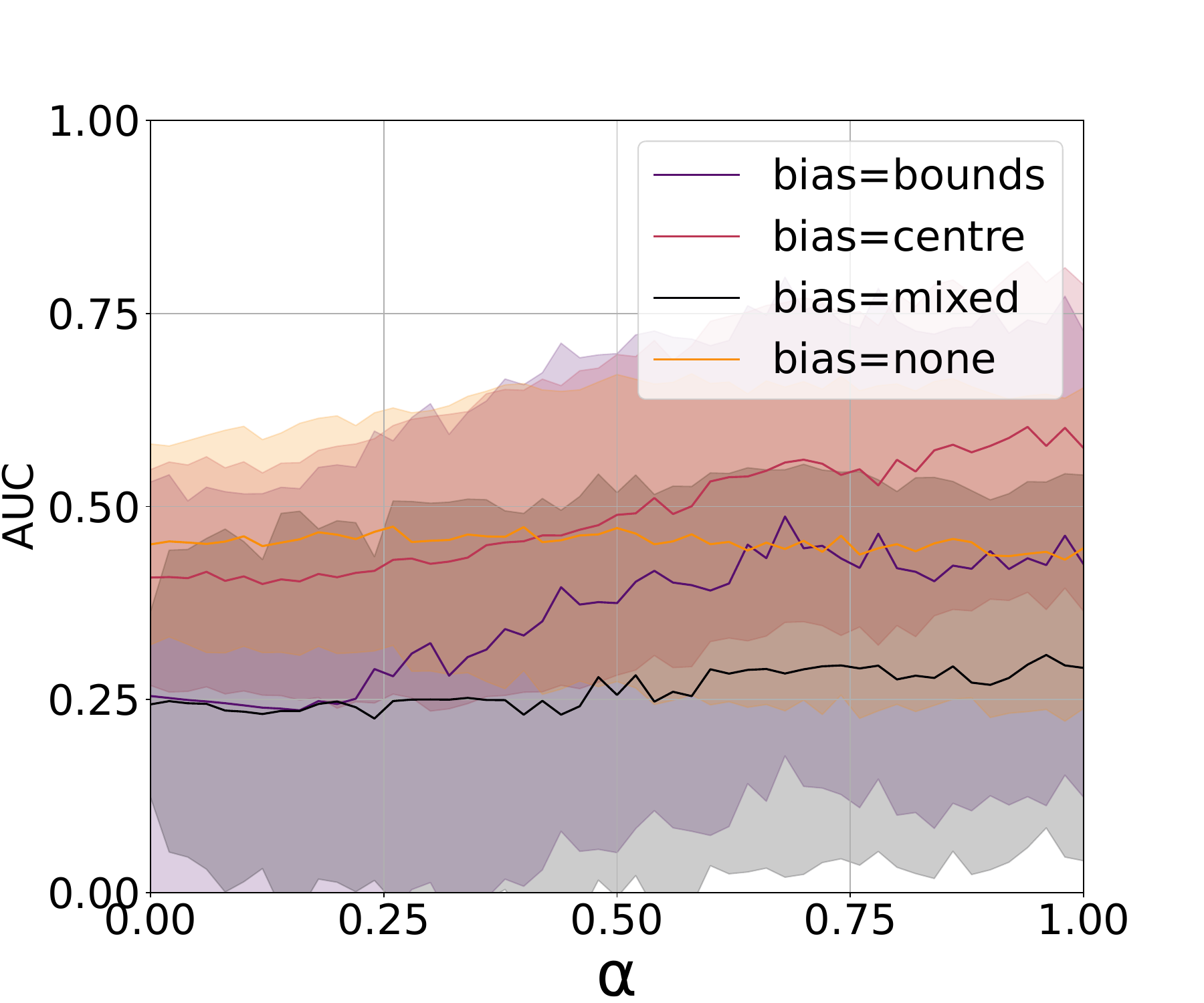}
        %\caption{Centre of bounds}
        \label{fig:gall-3}
    \end{subfigure}
    \begin{subfigure}[b]{0.24\textwidth}
        \includegraphics[height=2.6cm,trim=35mm 15mm 30mm 24mm,clip]{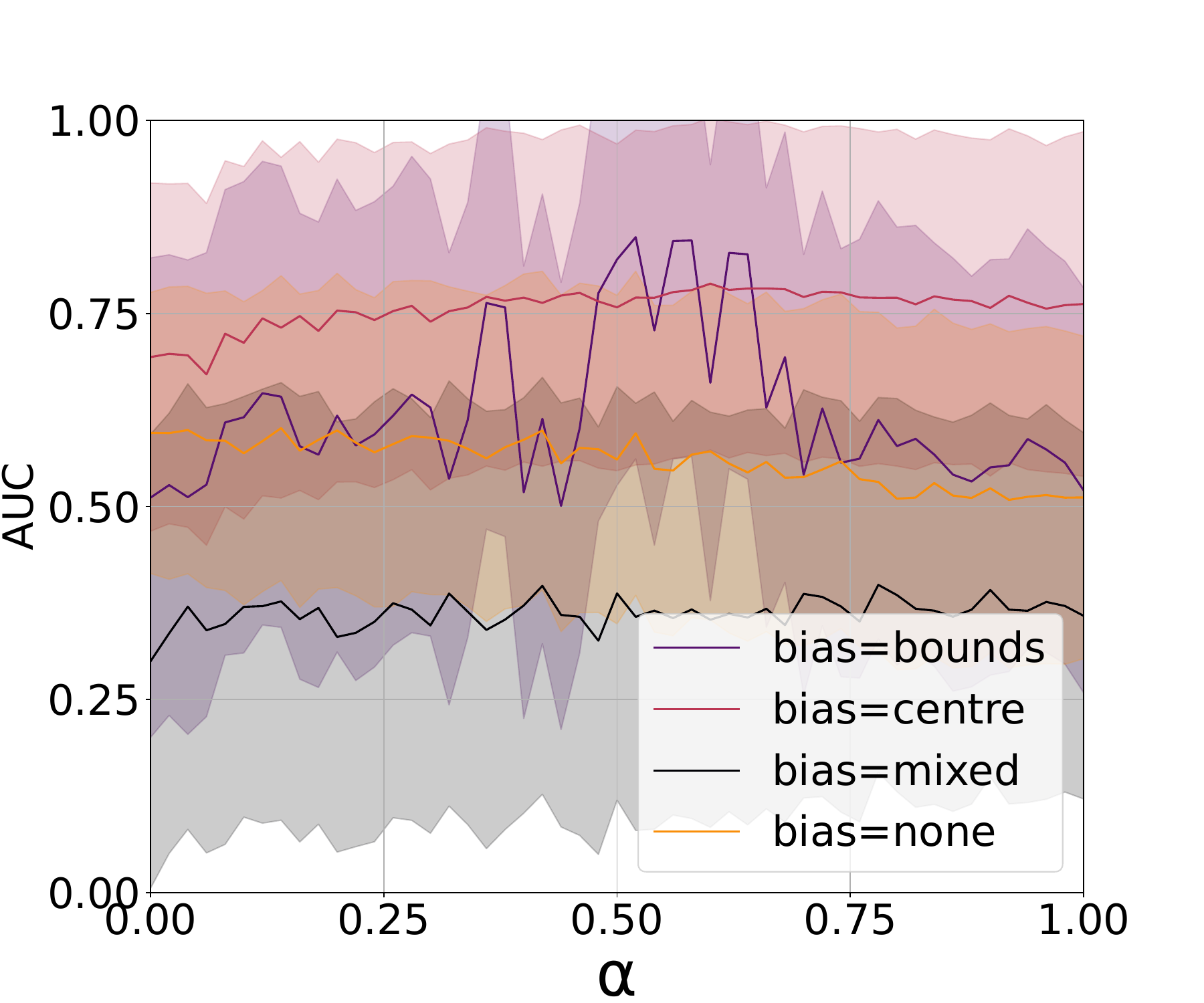}
        \label{fig:gall-4}
    \end{subfigure}
    \caption{Median and variance for AUC over 30 runs for CMA-ES variants split into bias classes across increasing \(\alpha\) for optimum placements (left to right) bounds, centre, centre of bounds, and random --- and for base BBOB function pairs (rows, top to bottom): \emph{f3}-\emph{f1}; \emph{f15}-\emph{f1}; \emph{f16}-\emph{f1}; and \emph{f21}-\emph{f1}.}
    \label{fig:performance}
\end{figure}

\section{Results}

%\subsection{Bias and performance}

Figure \ref{fig:function-viz} presents, for different base BBOB function pairings, states of affine combination for increasing \(\alpha\) (left to right), which is the proportion of the Sphere function \emph{f1}. Colour represents fitness and the global optimum is placed here near the centre. 

Notice from the left-most plots that with $\alpha=0$, the function contains no aspect of \emph{f1}. For Figure \ref{fig:function-viz} rows 1 and 2, their intermediate stages ($0.25 \leq \alpha \leq 0.75$) show the influence of the {\lq{roundness}\rq} from \emph{f1} being added into the function. In the case of Figure \ref{fig:function-viz} rows 3 and 4, the effect of increasing $\alpha$ manifests differently, with the bowl shape and concentric structure of \emph{f1} beginning to appear. We can see that when $\alpha=1$, the function is entirely \emph{f1}.   

Figure \ref{fig:performance} presents algorithm performance (AUC median and variance) over 30 runs for the top-scoring CMA-ES variants in each bias class across increasing \(\alpha\) (shown on the horizontal axis), split by location of the optimum (left to right: bounds, centre, centre of bounds, and random) and for base function pairs (top to bottom): \emph{f3}-\emph{f1}, \emph{f15}-\emph{f1}, \emph{f16}-\emph{f1}, and \emph{f21}-\emph{f1}.

The first three rows (that is, the first three function pairs) show similar patterns and trends. Generally, increasing $\alpha$ (proportion of \emph{f1}) is associated with increasing performance. Notice by comparing the median lines of, for example, the second plot of the first row with the other three plots in the same row that placing the optimum at the centre leads to the best performances by the algorithms, regardless of bias class. Algorithms perform at their worst when the optimum is placed at the bounds in at least one of the two co-ordinates (see the first and third columns of plotss). 

We now consider how algorithms from the different bias classes compare. Observe from the first and third plot in the first three rows that the \emph{centre} and \emph{none} classes of algorithms perform best (in that order) when the optimum is located near the bounds of the function. This is a curious result: intuitively, the \emph{bounds} algorithms would do the best. We checked some of the algorithm runs and noticed that \emph{centre} algorithms appear to have more freedom of movement --- making several small improvements in fitness. On the other hand, \emph{bounds} algorithms seem to sometimes get stuck in these cases when the optimum is on the bounds --- struggling to find a fitness improvement and advance towards the optimum location. We leave a statistical analysis of this phenomenon for future study. In the cases where the optimum is placed either centrally or randomly, the best-performing classes of algorithms are \emph{bounds} and \emph{centre} (notice the second and fourth plots in the first three rows). While it makes sense that \emph{centre} algorithms would perform best on these, the high performance of \emph{bounds} is less intuitive. From examination of a sample of performance runs, it seems that although \emph{bounds} algorithms may begin the search as biased towards the bounds, in the specific case where the optimum is centrally located they seem to be able to navigate towards the centre over the course of the search. It seems that the performance of \emph{bounds}-biased algorithms depends on the location of the optimum, but not in the way which might be expected: if the optimum is at the bounds, they may struggle; if it is at the centre, they do better. Note from the Figures that overall, the \emph{mixed} class of algorithms is the worst performing. 

The last function pairing, shown in the last row, differs from the other pairings substantially when the optimum is placed centrally (second column). We notice that performance is excellent (with AUC near to 1 in some cases) and that the trend with respect to $\alpha$ has reversed: increasing $\alpha$ is here associated with a decrease in performance. The explanation for this finding can probably be found in the nature of the original base BBOB function \emph{f21}, where the global optimum can be found near the bounds at the bottom of a half-funnel shape. Observe from the lowest-left plot in Figure \ref{fig:function-viz} that when the optimum is placed in the centre, this appears to stretch and mirror the half-funnel structure which leads to the optimum. We therefore speculate that this stretched funnel surrounding the optimum (in the case when it is centrally placed) is the reason for high algorithmic performance.

\section{Conclusions}

This study has systematically explored the interplay between the performance of configurations of the modular Covariance Matrix Adaptation Evolution Strategy (modCMA) and structural bias within different optimisation landscapes. Through the extensive configuration testing of modCMA, encompassing $435\,456$ configurations, we have shown that specific modules notably influence the algorithm's structural bias. Key insights include the significant impact of modules like covariance adaptation and elitism in modulating structural bias towards the centre of the search space and bound correction method Saturate towards the boundaries of the search space.

To investigate the effects of different forms of SB on algorithm performance, we generated pairwise affine recombinations of BBOB functions with varying proportions of each composite function. For each function we considered four strategies for placing the optimum. The configurations with highest confidence per SB class (predicted by the Deep-BIAS tool) of modCMA, were run $30$ times on the affine-combined functions. The results showed that when the optimum is placed at the centre, bounds-biased and centre-biased algorithms perform best. The reason for this is likely that when the optimum is near the centre, bounds-biased algorithms can navigate in the right direction even if they have an inherent bias towards the bounds --- the bounds structural bias is mainly caused by the bounds correction method \emph{saturate}, which does not often become active when the search leads away from the bounds. When the optimum is near the bounds, centre-biased and unbiased algorithms are performing better. We believe that centre-biased algorithms have more freedom of movement, and that bounds-biased algorithms with \emph{saturate} bound correction method become early stuck when the optimum is at the bounds. %Future work will further analyse this specific behaviour. 

Future research should focus on extending the analysis of structural bias effects into higher dimensional spaces. As dimensionality increases, the complex interplay between geometry of high-dimensional spaces, structural bias and landscape features may exhibit different characteristics that could bring additional insights. This future work will not only deepen our understanding of structural bias in iterative algorithms but also guide the development of more robust strategies for tackling complex optimisation problems.

\bibliographystyle{splncs03}
\bibliography{ppsn-paper}

\end{document}